\documentclass{article}
\usepackage[preprint,nonatbib]{corl_2026}
\usepackage[T1]{fontenc}
\usepackage{amsmath,amssymb,booktabs,multirow,enumitem,graphicx}
\usepackage{algorithm,algpseudocode,adjustbox,dsfont,float}
\usepackage[font=small,labelfont=bf,labelsep=period]{caption}
\usepackage[numbers,sort&compress,square]{natbib}
\usepackage[section]{placeins}
\newcommand{\modelA}{A2A-AffordGen}
\newcommand{\modelB}{A2A-GroundingModel}
\newcommand{\modelC}{A2A-Policy}
\newcommand\blfootnote[1]{\begingroup\renewcommand\thefootnote{}\footnotetext{#1}\addtocounter{footnote}{-1}\endgroup}
\hypersetup{pdftitle={Affordance2Action: Task-Conditioned Scene-level Affordance Grounding for Real-Time Manipulation},pdfauthor={Litao Liu, Yifan Han, Pengfei Yi, Wenbo Yu, Hanqing Wang, Haoran Du, Enze Yuan, Zilin Yuan, Ruiding Feng, Michael Liu, Qi Zhang, Jingjin Yu},pdfsubject={Extended arXiv version}}
\title{Affordance2Action: Task-Conditioned Scene-level\\Affordance Grounding for Real-Time Manipulation}
\author{\normalfont\small
Litao Liu$^{1,*}$, Yifan Han$^{1,*}$, Pengfei Yi$^{1}$, Wenbo Yu$^{1}$,\\
\normalfont\small Hanqing Wang$^{2}$, Haoran Du$^{1}$, Enze Yuan$^{1}$, Zilin Yuan$^{1}$,\\
\normalfont\small Ruiding Feng$^{1}$, Michael Liu$^{1}$, Qi Zhang$^{3}$, Jingjin Yu$^{1,\dagger}$\\
\normalfont\small $^{1}$Department of Computer Science, Rutgers University-New Brunswick\\
\normalfont\small $^{2}$The Hong Kong University of Science and Technology (GZ)\\
\normalfont\small $^{3}$Shanghai AI Laboratory\\
\normalfont\small $^{*}$Equal contribution.\quad $^{\dagger}$Corresponding author.\\
\normalfont\small \url{https://arc-l.github.io/a2a/}}
\begin{document}
\maketitle
\pagestyle{plain}\raggedbottom
\blfootnote{Corresponding author: Jingjin Yu (\texttt{jingjin.yu@cs.rutgers.edu}). Project leaders: Litao Liu (\texttt{litao.liu@rutgers.edu}) and Yifan Han (\texttt{hanyifan2024@ia.ac.cn}). This work was completed by Yifan Han, Pengfei Yi, and Wenbo Yu during their internship at Rutgers University-New Brunswick.}
\vspace{-10pt}
\begin{figure}[H]
\centering
\includegraphics[width=\linewidth]{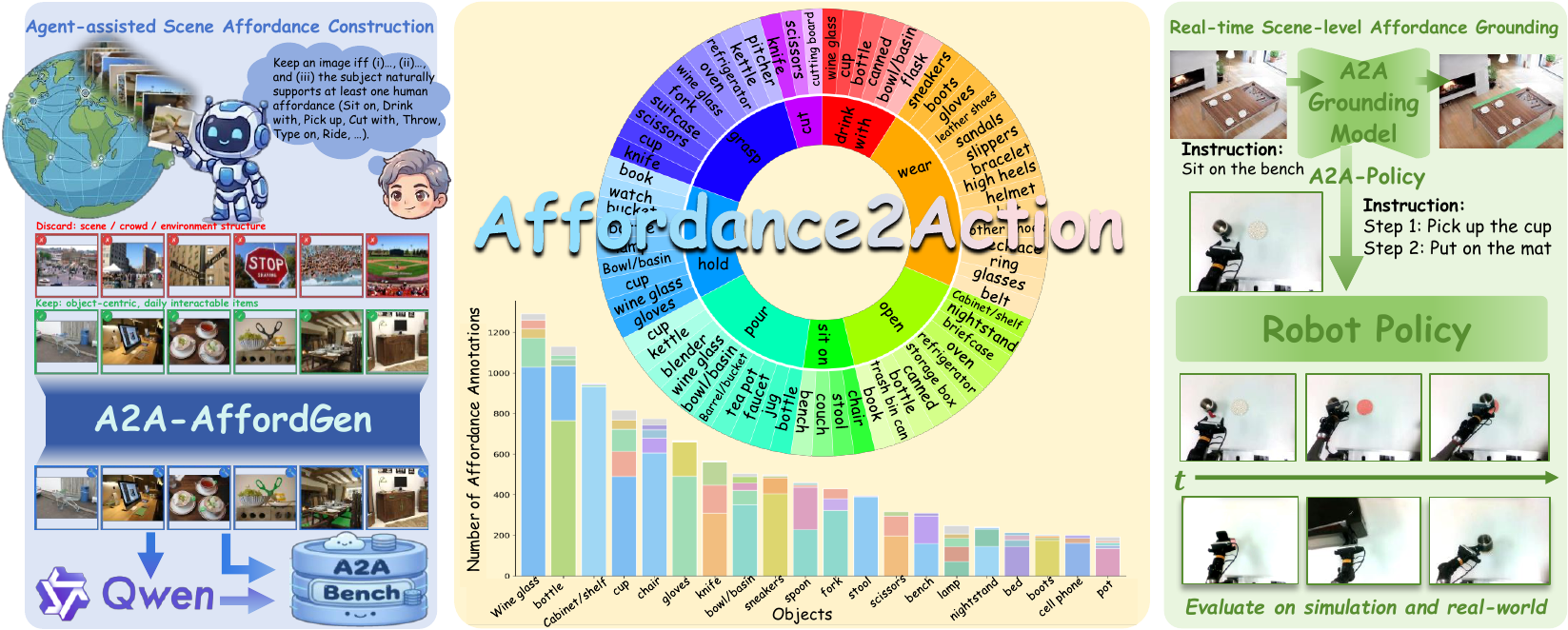}
\caption{\textbf{Overview of Affordance2Action (A2A).} A2A-AffordGen constructs A2A-Bench with task-conditioned functional-region annotations in natural multi-object scenes. The annotations train A2A-GroundingModel, whose real-time predictions supply spatial priors to A2A-Policy. The annotation teacher is separate from the affordance model used for robot control.}
\label{fig:a2a-teaser}
\end{figure}

\begingroup
\setlength{\leftmargini}{1em}
\begin{abstract}
Task-conditioned manipulation requires grounding instructions to task-relevant functional parts rather than object categories.
This setting is scene-dependent and often one-to-many in cluttered scenes: the same object may afford different interactions across tasks, while a single task may correspond to either one functional region or multiple valid functional regions, depending on the scene layout.
Existing affordance datasets and benchmarks remain misaligned with this setting, as they typically focus on grasping or object-level affordances, rely on synthetic scenes, or assume a single instruction-region correspondence.
We present \textbf{Affordance2Action (A2A)}, a benchmark-centered learning framework for scene-level, task-conditioned part affordance grounding.
At its core is \textbf{A2A-Bench}, a manipulation-oriented benchmark that covers both single-region and multi-region instruction correspondences in everyday scenes, with the latter highlighting the ambiguity and diversity of affordance grounding in realistic multi-object environments.
To construct it at scale, we build \textbf{A2A-AffordGen}, an agent-assisted annotation pipeline that combines language-model filtering, interactive part segmentation, instance-level mask-out refinement, task-reasoning instruction generation, and human verification.
A2A-Bench's supervision further supports diverse downstream applications, with real-time affordance grounding and affordance-conditioned manipulation policies as two representative examples.
Experiments show that A2A exposes substantial gaps in generic segmentation, VLM-based grounding, and affordance distillation baselines, while improving task-level localization and providing useful spatial priors for downstream manipulation.
\end{abstract}
\endgroup

\section{Introduction}

Language-guided manipulation grounds instructions in functional parts that guide action prediction~\citep{li2025coa}. For example, ``open the top drawer of the nightstand'' and ``place the cup to the nightstand'' target different parts of the same object. Grounding depends on task, scene layout, and strategy: instructions may admit multiple valid regions. Vision foundation models~\citep{li2025sam3, carion2025sam} segment explicitly described visual concepts, whereas manipulation requires inferring functional regions from task intent. Vision-Language-Action (VLA) policies learn these regions implicitly, reducing interpretability and adaptability. Explicit affordances provide a representation linking language grounding to action prediction.

RAGNet~\citep{wu2025ragnet} emphasizes grasping; InstructPart~\citep{wan2025instructpart} links language to parts but lacks object diversity, scene complexity, and scale. UAD~\citep{tang2025uad} reduces annotation costs by distilling foundation-model affordances, but simplified synthetic objects limit robustness in clutter. Most affordance formulations assume one-to-one instruction-region correspondence. Evaluating scene-level understanding instead requires one-to-many annotations covering all plausible task-relevant regions, not only the most salient region.

We propose \textbf{Affordance2Action (A2A)} for real-time, task-conditioned part grounding. Its \textbf{A2A-Bench} benchmark covers single- and multi-region correspondences in clutter. We filter large-scale natural images~\citep{shao2019objects365} for manipulation-relevant affordances and annotate functional regions through interactive segmentation~\citep{zhu2025segagent} with iterative mask-out refinement. These annotations represent interaction opportunities rather than embodiment-specific execution guarantees, serving as grounding targets and policy priors. Building on this supervision, \textbf{A2A-GroundingModel} adapts SAM3~\citep{carion2025sam} to predict affordance masks from image-instruction pairs without inference-time spatial prompts. Staged instruction adaptation and text-conditioned visual prompts connect explicit part descriptions to task intent. \textbf{A2A-Policy} uses these masks as priors for manipulation~\citep{chi2025diffusion}. Our contributions are:

\begin{itemize}[leftmargin=4mm,topsep=2pt,partopsep=0pt]
\setlength{\itemsep}{1pt}
\setlength{\parskip}{0pt}
    \item \textbf{A2A-Bench}: a scene-level benchmark linking task instructions to one or multiple functional regions.
    \item \textbf{A2A-AffordGen}: a pipeline for lower-cost labeling through filtering, interactive segmentation, mask-out refinement, instruction generation, and verification.
    \item  \textbf{A2A-GroundingModel} and \textbf{A2A-Policy}: real-time part grounding and manipulation models that turn benchmark supervision into spatial action priors.
\end{itemize}

\section{Related Work}
\label{sec:related_works}

\noindent\textbf{Affordance Learning.}
Affordance learning connects perception and manipulation through 2D masks~\citep{tang2025uad, li2023locate}, heatmaps~\citep{li2024one,luo2022learning}, or 3D surface priors from point clouds~\citep{IAGNet,deng20213d}. HOI-based approaches add temporal cues from pre-contact motion, contact regions, and post-contact dynamics~\citep{wang2026videoafford,bahl2023affordances}. Foundation models and MLLMs infer functional regions from multimodal inputs~\citep{yu2025seqafford,zhang2025a4,qian2024affordancellm,zhang2026openhoi}, with Affordance-R1~\citep{wang2026affordance} exploring deep affordance reasoning. Most prior work is object-centric or assumes one-to-one correspondence; A2A-Bench covers single- and multi-region grounding in real-world multi-object scenes.

\noindent\textbf{Affordance for Robot Policies.} Language and vision-language models support affordance-aware action reasoning. CoA-VLA~\citep{li2025coa} decomposes manipulation into object, grasp, spatial, and movement affordances to guide VLA policies. Affordance often remains external to the learned perception-action representation. RAGNet~\citep{wu2025ragnet} guides grasping with predicted affordance regions. Closest to our work, UAD~\citep{tang2025uad} distills foundation-model affordances into heatmaps for imitation learning. We study both mask highlighting and feature injection as task-conditioned action priors.

\noindent\textbf{Predictive and Generative Policy Learning.}
Related policy-learning approaches introduce additional structure through prediction, optimization, and data augmentation. FoAM~\citep{liu2026foam} combines multimodal goal conditions with foresight about the visual consequences of actions in multi-task imitation learning. RoboFlow4D~\citep{lin2026roboflow4d} predicts multi-frame 3D flows from observations and instructions to guide action generation. DyGRO-VLA~\citep{lin2026dygro} studies cross-task VLA optimization through shared representations and grouped residual updates. BiTrajDiff~\citep{qing2025bitrajdiff} uses forward and backward diffusion of trajectories to augment offline reinforcement-learning data. Our study focuses on functional-part grounding as an explicit intermediate representation and evaluates how masks and region-conditioned features support downstream manipulation.

\section{Methodology}
\label{sec:methods}
\begin{figure}[htbp]
\centering
\includegraphics[width=1.0\linewidth]{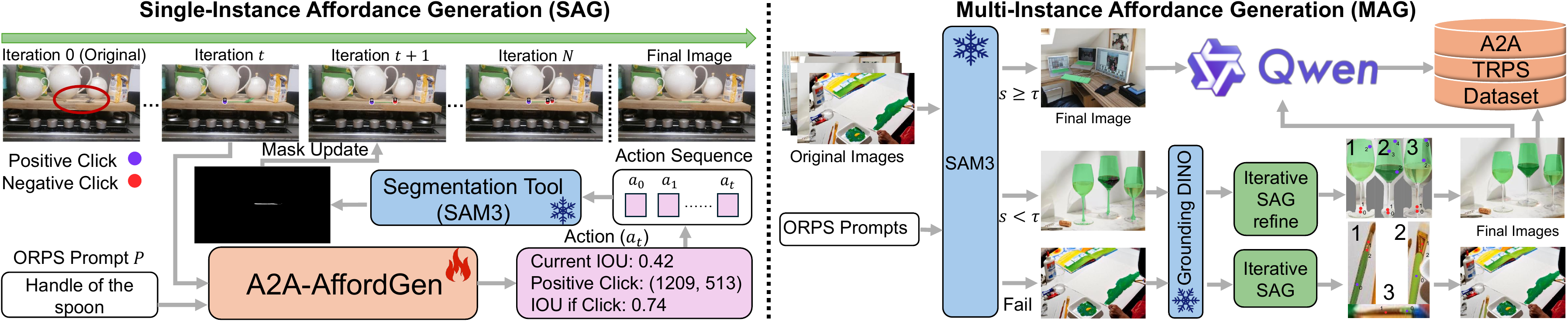}
\caption{\textbf{A2A-AffordGen.} \textbf{SAG (left)} predicts $\{+,-\}$ clicks from an ORPS prompt and current mask for frozen SAM3. \textbf{MAG (right)} triages SAM3 text-prompted masks, refines low-confidence instances with SAG, and reassembles scene-level multi-instance, multi-part annotations for A2A--TRPS.}
\label{fig:a2a-pipeline}
\end{figure}

\subsection{A2A-Bench}
\label{sec:a2a_bench}

One-to-many affordance annotation must identify task-relevant parts across multiple interactable objects in a scene.
Although open-vocabulary segmenters such as SAM3~\citep{carion2025sam} provide strong scene-level grounding, they struggle with small or visually subtle functional parts.
We introduce \modelA{}, a dual-regime annotation pipeline for constructing A2A-Bench (Fig.~\ref{fig:a2a-pipeline}), supporting both single-instance and multi-instance affordance generation.
Single-Instance Affordance Generation (SAG) refines isolated-object part masks through iterative point prompting; Multi-Instance Affordance Generation (MAG) extends it to cluttered multi-object scenes.
Both regimes follow the Oracle Referring Part Segmentation (ORPS) protocol~\citep{wan2025instructpart}, enabling part-level segmentation from general referring prompts.
We further use a VLM to author diverse Task Reasoning Part Segmentation (TRPS) instructions~\citep{wan2025instructpart} from the resulting masks, yielding task-conditioned data for downstream affordance model training (Sec.~\ref{sec:a2a_affordance_model}).

\textbf{Click-policy training.}
We first train the click policy shared by SAG and MAG. Inspired by SegAgent~\citep{zhu2025segagent}, we model point prompting of frozen SAM3 as a Markov decision process. Given crop $I$, ORPS prompt $P$, and mask $M_t$, state $s_t$ contains the mask overlay and prompt. Action $a_t=(p_t,u_t)\in\{+,-\}\times\Omega$ specifies click polarity and pixel; the transition is $M_{t+1}=\Phi(I,P,M_t,a_t)$, where $\Phi$ is frozen SAM3 and $\Omega$ the pixel grid.

We imitate SimpleClick's distance-maximizing oracle~\citep{liu2023simpleclick,xu2016deep}. For reference mask $M^\star$, false negatives and positives are $E_t^+=M^\star\setminus M_t$ and $E_t^-=M_t\setminus M^\star$. The oracle selects the dominant polarity and its chamfer-center:
\begin{equation}
p_t^\star = \operatorname*{arg\,max}_{p\in\{+,-\}}|E_t^p|,\quad
u_t^\star = \operatorname*{arg\,max}_{u\in E_t^{p_t^\star}}\mathrm{dist}(u,\partial E_t^{p_t^\star}),
\label{eq:sag-oracle}
\end{equation}
where $\mathrm{dist}$ is Euclidean distance to the error boundary. Oracle rollouts start at $M_0=\emptyset$ and stop at reference IoU $\kappa$ or step cap $T_{\max}$. Each action $a_t^\star=(p_t^\star,u_t^\star)$ provides next-mask quality and gain targets:
\begin{equation}
\begin{aligned}
v_t^\star &= \mathrm{IoU}\!\left(\Phi(I,P,M_t,a_t^\star),M^\star\right),\\
\delta_t^\star &= v_t^\star-\mathrm{IoU}(M_t,M^\star).
\end{aligned}
\label{eq:sag-quality}
\end{equation}
We adapt Qwen3.5-9B~\citep{team2026qwen3} with LoRA~\citep{hu2021lora} to emit $(p_t,u_t,\hat v_t,\hat\delta_t)$ as structured text. Token-level cross-entropy supervises the serialized oracle target $y_t^\star=(p_t^\star,u_t^\star,v_t^\star,\delta_t^\star)$:
\begin{equation}
\mathcal{L}_{\mathrm{agent}}=\mathrm{CE}\!\left(\pi_\theta(\cdot\mid s_t),y_t^\star\right).
\label{eq:sag-loss}
\end{equation}
The dominant-polarity rule chooses whether omitted or spurious pixels need correction. Selecting a pixel far from the error boundary places the oracle click inside a substantial error region rather than on an uncertain edge. The quality and gain targets are measured after the corresponding SAM3 transition, so they describe the proposed correction rather than the current mask. SAM3 only generates the next state; no pixel-level loss updates the VLM. This division of labor lets the language model choose where and how to correct a mask while the segmenter supplies its pixel geometry. Conditioning each action on the current mask is essential: the appropriate correction depends on the residual error after earlier interactions, rather than on the image and instruction alone.

\textbf{Single-instance generation (SAG).}
SAG uses the trained policy to refine a part mask for an object crop and ORPS prompt. Starting from an empty or supplied seed mask, it queries $\pi_\theta$ for a click and predicted next-mask quality and gain. Positive clicks address omitted target regions, and negative clicks suppress incorrectly included regions. 
At inference, reference masks are unavailable. SAG uses three stopping conditions:
(i) Low predicted gain. It stops before clicking if $\hat\delta_t<\Delta_{\min}$.
(ii) Target IoU reached. If the gain check passes and predicted next-mask IoU $\hat v_t>\tau_{\mathrm{iou}}$, SAG applies the click through frozen SAM3 and then stops.
(iii) Click budget exhausted. It stops after $T_{\max}$ clicks.
Alg.~\ref{alg:affordgen-inference} specifies the stopping test and mask update order.

\textbf{Multi-instance generation (MAG).}
For scene $I$ and prompt $P$, SAM3 text prompting supplies candidate masks $(m_i,s_i)$, while Grounding DINO~\citep{liu2024grounding} supplies object boxes. MAG routes instances through three branches:
(i) Direct accept ($s_i\geq\tau$). Accept the confident SAM3 mask as-is.
(ii) SAG-refine ($s_i<\tau$). Use the low-confidence mask as the seed for SAG within its matched Grounding DINO crop.
(iii) SAG-from-scratch (no matched mask). Run SAG from $M_0=\emptyset$ in each unmatched Grounding DINO box, including scenes where SAM3 finds no mask.
For seeded refinement, MAG dilates $m_i$ by kernel $\mathcal K$~\citep{haralick1987image}, fills pixels outside $m_i\oplus\mathcal K$ with neutral gray, and crops the result and seed mask to its box. Inverse crop mapping returns refined masks $\tilde M_i$ to scene coordinates; direct acceptance uses the identity mapping. The scene-level output is
\begin{equation}
\mathcal M_P(I)=\bigsqcup_i\mathcal C_i^{-1}(\tilde M_i),
\label{eq:mag-reassembly}
\end{equation}
where $\mathcal C_i^{-1}$ reverses the corresponding crop transform. Retaining individual masks supports one-to-many supervision rather than reducing a scene to one target. MAG reuses SAG's single-instance click policy without retraining. Qwen3-VL-32B-Instruct~\citep{bai2025qwen3vl} then generates diverse TRPS instructions for task-conditioned grounding. The crop wrapper lets the same single-instance policy operate in clutter without changing its training objective. Dilation preserves a narrow context band around the part, while gray mask-out suppresses neighboring objects that could attract subsequent clicks. Confident predictions bypass refinement; ambiguous and missing instances receive additional interaction.

\begin{algorithm}[htbp]
\small
\algtext*{EndIf}
\algtext*{EndFor}
\algtext*{EndFunction}
\caption{A2A-AffordGen inference: SAG and MAG.}
\label{alg:affordgen-inference}
\begin{algorithmic}[1]
\Require image $I$, prompt $P$, policy $\pi_\theta$, SAM3 $\Phi$, detector $G$, thresholds $\tau,\tau_{\mathrm{iou}},\Delta_{\min}$, kernel $\mathcal K$, cap $T_{\max}$
\Function{SAG}{$I,P,M$}
  \For{$t=0,\ldots,T_{\max}-1$}
    \State $(p,u,\hat v,\hat\delta)\gets\pi_\theta(I,P,M)$
    \If{$\hat\delta<\Delta_{\min}$}
      \State \textbf{break}
    \EndIf
    \State $M\gets\Phi(I,P,M,(p,u))$
    \If{$\hat v>\tau_{\mathrm{iou}}$}
      \State \textbf{break}
    \EndIf
  \EndFor
  \State \Return $M$
\EndFunction
\Function{MAG}{$I,P$}
  \State $\{(m_i,s_i)\}\gets\Phi_{\mathrm{text}}(I,P)$; $\mathcal B\gets G(I,P)$
  \State $\mathcal M\gets\emptyset$
  \For{each $(m_i,s_i)$}
    \If{$s_i\geq\tau$}
      \State $\mathcal M\gets\mathcal M\cup\{m_i\}$
    \Else
      \State $b\gets$ box in $\mathcal B$ matched to $m_i$
      \State $I_b\gets$ crop $b$ after gray mask-out outside $m_i\oplus\mathcal K$
      \State $\tilde M\gets\Call{SAG}{I_b,P,\mathrm{Crop}(m_i,b)}$
      \State $\mathcal M\gets\mathcal M\cup\{\mathcal C_b^{-1}(\tilde M)\}$
    \EndIf
  \EndFor
  \For{each unmatched box $b\in\mathcal B$}
    \State $\tilde M\gets\Call{SAG}{\mathrm{Crop}(I,b),P,\emptyset}$
    \State $\mathcal M\gets\mathcal M\cup\{\mathcal C_b^{-1}(\tilde M)\}$
  \EndFor
  \State \Return $\mathcal M$
\EndFunction
\end{algorithmic}
\end{algorithm}

\textbf{A2A-Bench composition and scale.}
A2A-Bench is built from large-scale in-the-wild images in three stages (Fig.~\ref{fig:a2a-teaser}, left).
(i) Point-prompting policy training corpus. We curate $\sim$40K single-object part masks from seven sources, including RAGNet, HANDAL, and InstructPart~\citep{wu2025ragnet, guo2023handal, wan2024instructpart, fang2020graspnet, o2024open, zhu2023egoobjects, qian2023understanding}, and normalize them into a consistent ORPS format.
(ii) Human-verified scene core. From Objects365~\citep{shao2019objects365}, we first use VLM-based filtering to discard unsuitable images, such as pure scenes, crowds, or environment-structure images, and retain object-centric daily interactable items. We then manually annotate 5{,}000 multi-object multi-part scenes spanning diverse tabletop and household categories, including $\sim$18K single-part masks. The masks from (i) and the cropped masks from (ii) are expanded via the SimpleClick distance-maximizing oracle~\citep{liu2023simpleclick} into $\sim$150K click-trajectory samples used to fine-tune \modelA{}.
(iii) Automatic scene-level scaling.
The trained \modelA{} runs MAG over the unseen Objects365 dataset to auto-generate scene-level annotations.
We release an initial batch of 5{,}000 generated multi-instance scenes after manual quality verification, reducing full manual mask annotation to lightweight human review.
The full constructed corpus is used for training; separately annotated test scenes remain held out.

\textbf{Annotation criteria and one-to-many structure.}
A valid region corresponds to a functional object part for which contact advances the instruction and is reachable in the depicted scene. This criterion identifies a visual interaction opportunity; it does not guarantee reachability for every robot embodiment or a successful action trajectory. Six trained annotators receive disjoint assignments, and a reviewer checks the annotations for consistency. In the initial 5,000-scene release, 77.2\% of scenes contain at least two valid regions. This prevalence motivates retaining separate masks during MAG reassembly and evaluating both instance coverage and count error.

\textbf{Separation of annotation and deployment.}
The click policy uses reference masks to construct training trajectories, but receives no reference mask at inference. Likewise, human verification improves the released supervision rather than adding privileged information to the online robot policy. A2A-AffordGen can spend multiple SAM3 interactions on an annotation because it operates during dataset construction. A2A-GroundingModel learns from the resulting supervision and predicts task-conditioned regions without that iterative annotation loop. This teacher--student separation connects annotation quality to deployment efficiency while keeping the two evaluation settings distinct.

The serialized training targets, stopping order, and crop-coordinate handling are detailed in App.~\ref{app:affordgen-details}.

\subsection{A2A-GroundingModel}

\label{sec:a2a_affordance_model}
\begin{figure}[htbp]
\centering
\includegraphics[width=1.0\linewidth]{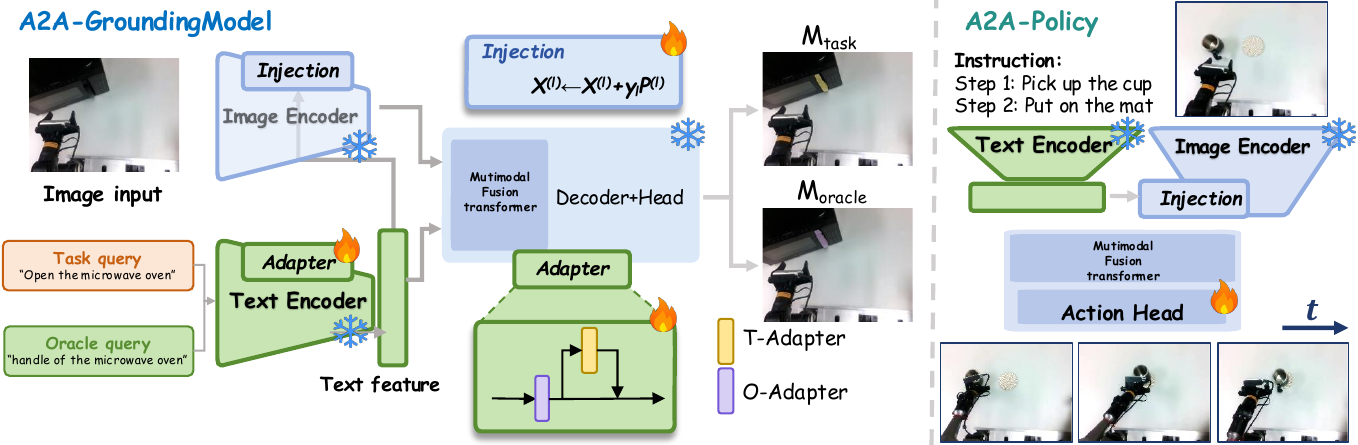}
\caption{\textbf{A2A-GroundingModel and A2A-Policy.}
Lightweight adapters and text-conditioned visual prompts adapt frozen SAM3 for affordance grounding. Explicit mask highlighting or implicit feature injection transfers this prior to manipulation.}
\label{fig:a2a-policy}

\end{figure}

\modelB{} learns real-time functional-region grounding from A2A-Bench supervision (Fig.~\ref{fig:a2a-policy}). Given image $I$ and instruction $q$, it predicts functional-part masks $\hat M$ indicating task-relevant interaction regions. Task-conditioned grounding must infer the part implied by manipulation intent: ``open the microwave oven'' targets its handle rather than the entire appliance. ORPS queries explicitly name the part ``handle of the microwave oven'', whereas TRPS queries describe the intended action. Paired ORPS/TRPS supervision connects explicit part localization to task reasoning while retaining the same reference regions.

\textbf{Staged instruction adaptation.}
Inspired by SAM3-I~\citep{li2025sam3}, we insert lightweight bottleneck adapters into selected language, fusion, decoder, and mask prediction blocks of frozen SAM3. At block $\ell$, hidden state $\mathbf x^{(\ell)}$ follows an oracle route $A_o$ (O-Adapter), or a bridge $A_{o\to t}$ followed by task adapter $A_t$ (T-Adapter):
\begin{equation}
\begin{aligned}
\mathbf x_o^{(\ell)} &= \mathbf x^{(\ell)}+A_o^{(\ell)}\!\left(\mathrm{LN}(\mathbf x^{(\ell)})\right),\\
\mathbf r^{(\ell)} &= \mathbf x^{(\ell)}+A_{o\to t}^{(\ell)}\!\left(\mathrm{LN}(\mathbf x^{(\ell)})\right),\\
\mathbf x_t^{(\ell)} &= \mathbf r^{(\ell)}+A_t^{(\ell)}\!\left(\mathrm{LN}(\mathbf r^{(\ell)})\right).
\end{aligned}
\label{eq:gm-adapters}
\end{equation}
Here $\mathrm{LN}$ denotes layer normalization; $o$ and $t$ denote Oracle query and Task query in Fig.~\ref{fig:a2a-policy}. Connected routes reduce interference between part recognition and task reasoning; inference uses the task route. The distinction between query types separates two demands on the model. An oracle query supplies the functional-part name, whereas a task query requires selecting that part from the intended action before localizing it. Because the paired queries refer to the same regions, the explicit query provides a localization anchor during adaptation. Deployment uses task instructions without part names, reference masks, or oracle-route predictions.

\textbf{Text-conditioned visual prompts.}
Let $\mathbf T\in\mathbb R^{L\times d_t}$ be language tokens and $\mathcal I_q$ their valid, non-padding positions. Masked averaging gives instruction vector $\mathbf z\in\mathbb R^{d_t}$. Patch projection and a shallow visual network produce image tokens $\mathbf V\in\mathbb R^{HW\times d_p}$. A learned text bias $\mathbf b$ and channel gate $\mathbf g$ modulate these tokens using spatial weights $\alpha_j$:
\begin{align}
&\mathbf z=\frac{1}{|\mathcal I_q|}\sum_{i\in\mathcal I_q}\mathbf T_i,\quad
\mathbf b=W_b\mathrm{LN}(\mathbf z),\notag\\
&\mathbf g=\sigma\!\left(W_g\mathrm{LN}(\mathbf z)\right),\quad
\alpha_j=\frac{\exp(\mathbf V_j^\top\mathbf b/\sqrt{d_p})}
{\sum_k\exp(\mathbf V_k^\top\mathbf b/\sqrt{d_p})},\label{eq:gm-prompts}\\
&\mathbf S_j=\mathrm{LN}\!\left(\mathbf V_j\odot(1+\alpha_j\mathbf g)+\mathbf b\right),\quad
\mathbf X^{(\ell)}\leftarrow\mathbf X^{(\ell)}+\gamma_\ell\mathbf P^{(\ell)}.\notag
\end{align}
Here $\sigma$ is sigmoid and $\odot$ element-wise multiplication; the weights $\alpha_j$ normalize over spatial locations. Spatial weighting identifies relevant locations, while the gate modulates feature channels. Layer-specific projections map seed $\mathbf S$ to prompts $\mathbf P^{(\ell)}$ matching visual features $\mathbf X^{(\ell)}$; $\gamma_\ell$ is learned. Prompts enter later visual encoder blocks, where semantically structured features support task conditioning. This introduces manipulation intent during visual feature formation, emphasizing small or weakly textured parts that whole-object semantics can obscure. Adapters modify hidden-state processing, while visual prompts add instruction-dependent signals to image tokens. Together they produce task-specific image representations while keeping the backbone frozen.

\textbf{Training objective and schedule.}
Hungarian assignment matches predictions to reference masks. The segmentation objective combines classification, box regression, generalized-IoU, and mask losses:
\begin{equation}
\mathcal L_{\mathrm{seg}}=\lambda_{\mathrm{cls}}\mathcal L_{\mathrm{cls}}
+\lambda_{\mathrm{box}}\mathcal L_{\mathrm{box}}
+\lambda_{\mathrm{giou}}\mathcal L_{\mathrm{giou}}
+\lambda_{\mathrm{mask}}\mathcal L_{\mathrm{mask}}.
\label{eq:gm-seg}
\end{equation}
For paired ORPS/TRPS queries, spatial distributions $\mathbf P_o$ and $\mathbf P_t$ from the oracle and task routes are aligned by a bidirectional consistency objective:
\begin{equation}
\mathcal L_{\mathrm{KL}}=\tfrac12\!\left[D_{\mathrm{KL}}(\mathbf P_o\Vert\mathbf P_t)
+D_{\mathrm{KL}}(\mathbf P_t\Vert\mathbf P_o)\right].
\label{eq:gm-kl}
\end{equation}
The task-query term supervises $\hat M_{\mathrm{task}}$, corresponding to $M_{\mathrm{task}}$ in Fig.~\ref{fig:a2a-policy}. For reference $M$ and oracle--task disagreement weight $\mathbf W_{\mathrm{task}}$ on grid $\Omega$,
\begin{equation}
\mathcal L_{\mathrm{task}}=\frac{1}{|\Omega|}\sum_{u\in\Omega}
\mathbf W_{\mathrm{task}}(u)\,\mathcal L_{\mathrm{mask}}\!\left(\hat M_{\mathrm{task}}(u),M(u)\right).
\label{eq:gm-task}
\end{equation}
The combined objective is
\begin{equation}
\mathcal L=\mathcal L_{\mathrm{seg}}+\lambda_{\mathrm{KL}}\mathcal L_{\mathrm{KL}}
+\lambda_{\mathrm{task}}\mathcal L_{\mathrm{task}}.
\label{eq:gm-total}
\end{equation}
Stage I learns localization from explicit ORPS part descriptions. Stage II transfers this ability to TRPS queries that imply the functional part through task intent. Stage III jointly trains the paired routes with consistency regularization to preserve part localization while aligning task-conditioned predictions. Only adapters and visual prompt modules update; the SAM3 backbone remains frozen. The oracle route supplies training supervision, while inference requires only $I$ and the task instruction.

The step-by-step adapter, visual-prompt, and training formulations are provided in App.~\ref{app:grounding_model_details}.

\subsection{A2A-Policy}
\label{sec:a2a_policy_learning}

We instantiate \modelB{} (Sec.~\ref{sec:a2a_affordance_model}) in two manipulation-policy variants.
Both variants share an action-chunking diffusion head~\citep{chi2025diffusion, zhao2023learning} and use different affordance interfaces.
\modelB{} remains frozen.

\textbf{Explicit mask highlighting.}
In the explicit variant, inspired by UAD~\citep{tang2025uad}, \modelB{} grounds task-relevant functional regions on the raw RGB observation. For policy conditioning, we use the highest-confidence grounded region as a colored alpha overlay before feeding the observation to the action head. The remaining policy matches the diffusion baseline, isolating the highlighting effect.

\textbf{Implicit feature injection.}
In the implicit variant (Fig.~\ref{fig:a2a-policy}, right), the frozen \modelB{} acts as the visual-text encoder. We process each frame with text-conditioned visual prompt injection (Sec.~\ref{sec:a2a_affordance_model}). We select the highest-confidence predicted region and threshold its mask before pooling the multi-scale FPN features~\citep{fang2025sam2act}, following Afford-VLA~\citep{wang2026afford}. Region-conditioned tokens are projected to the policy feature width. A shared multimodal transformer fuses them with adapted instruction and state tokens; only its $\langle\mathrm{ACT}\rangle$ token conditions the diffusion head. This replaces unfiltered coarse feature pooling while keeping the training setup unchanged. Affordance guides feature formation and spatial aggregation without overlays. 

The two variants test where to introduce the same grounding prior: explicit conditioning modifies the visual observation, whereas implicit conditioning restricts the features passed to the policy. Freezing the grounding model keeps policy learning from changing its masks to fit individual demonstrations. Selecting one region is a deliberate interface simplification for these manipulation tasks; it does not evaluate a planner that reasons over or acts on all regions produced by the scene-level model.

\section{Experiments}
\label{Experiment}
We evaluate the framework at three levels: annotation quality from \modelA{}, task-conditioned prediction from \modelB{}, and downstream manipulation with \modelC{}. These evaluations use different inputs and outputs. ORPS supplies an explicit part description to assess mask generation, TRPS asks a model to infer the relevant part from a task instruction, and policy evaluation measures whether the robot completes that task. We therefore report each protocol separately rather than treating a high grounding score as a substitute for successful execution.

\subsection{Benchmarks, Metrics, and Implementation}
\label{sec:exp-setup}

\textbf{Grounding benchmarks.}
We evaluate \modelA{} on separately annotated, held-out ORPS sets: 200 single-instance SAG scenes and 200 multi-instance MAG scenes. The same examples are used across seeds 0, 1, and 2, and the test images are excluded from the constructed training corpus. We also evaluate \modelB{} on these ORPS sets to compare the annotation teacher with the real-time student under the same query type. Separately, online task-conditioned grounding is evaluated on 200 fixed TRPS mask--query pairs, with mask accuracy, latency (ms/image), and throughput (FPS). Zero-shot transfer uses RefCOCO, RefCOCO+, and RefCOCOg~\citep{yu2016modeling,kazemzadeh2014referitgame,mao2016generation}, reporting referring-segmentation cIoU without benchmark-specific fine-tuning.

\textbf{Mask-overlap and threshold metrics.}
For an evaluated prediction--reference pair $(\hat M_n,M_n^\star)$, let $I_n=|\hat M_n\cap M_n^\star|$, $U_n=|\hat M_n\cup M_n^\star|$, and $J_n=I_n/U_n$. We report
\begin{equation}
\mathrm{gIoU}=\frac{1}{N}\sum_{n=1}^{N}J_n,\qquad
\mathrm{cIoU}=\frac{\sum_{n=1}^{N}I_n}{\sum_{n=1}^{N}U_n}.
\label{eq:overlap-metrics}
\end{equation}
Here gIoU is the mean of the evaluated mask-pair IoUs, not generalized box IoU. It weights mask pairs equally, whereas cIoU pools intersection and union pixels and therefore gives larger unions more weight. Define $P(t)=N^{-1}\sum_n\mathbf{1}[J_n\geq t]$. P@50 is $P(0.50)$ and P@50--95 is the mean of $P(t)$ over the ten thresholds $t=0.50,0.55,\ldots,0.95$. Higher thresholds require more precise masks. Grounding accuracy values are computed per run and averaged over the three seeds; all overlap and threshold-based scores are presented as percentages.

\textbf{Multi-instance set metrics.}
Pixel overlap alone does not reveal whether a method has recovered every valid instance. Let $K_p$ and $K_g$ be the numbers of predicted and reference masks in a scene, and set $K=\max(K_p,K_g)$. We pad the smaller set with empty masks and give padded pairs zero IoU. The set-overlap score is
\begin{equation}
\mathrm{sIoU}=\max_{\sigma\in\mathfrak S_K}\frac{1}{K}
\sum_{k=1}^{K}\mathrm{IoU}\!\left(\hat M^{(k)},M^{\star(\sigma(k))}\right),
\label{eq:set-iou}
\end{equation}
where $\mathfrak S_K$ contains the permutations and Hungarian matching finds the optimum. One prediction can match at most one reference; padding penalizes missed and extra regions. If $T_n$ real matched pairs have IoU at least 0.5 in scene $n$, set recall is $R_n=T_n/K_{g,n}$ (Set-R), and set precision is $P_n=T_n/K_{p,n}$ (Set-P); padded pairs never count as successful matches. Count MAE is $N^{-1}\sum_n|K_{p,n}-K_{g,n}|$. Set-R measures instance coverage, Set-P measures the fraction of useful proposals, and Count MAE measures count error irrespective of boundary shape. These measures complement rather than replace pixel overlap. The reported set scores are percentages, while Count MAE is an instance count with lower values better.

\textbf{Policy benchmarks and success metric.}
Policy evaluation uses task completion rather than segmentation overlap. Each of ten LIBERO-Object tasks~\citep{liu2023libero} has 100 expert demonstrations and 50 evaluation episodes; we report mean task success rate. Four real-world tasks on an AgileX Piper six-DoF arm each use 100 teleoperated demonstrations and ten evaluation trials with wrist/head RGB observations. The real-robot aggregate is the total number of successful trials out of 40. These counts are kept separate from the three-seed grounding averages.

\textbf{Grounding implementation.}
All training and deployment use NVIDIA RTX PRO 6000 Blackwell. \modelA{} fine-tunes Qwen3.5-9B with LoRA and permits at most eight clicks per instance. Frozen SAM3 converts the predicted clicks into masks. \modelB{} trains adapters and visual prompts within frozen SAM3, with the task route used at deployment. \modelA{} generates supervision; the real-time \modelB{} supplies the affordance signal to both policy variants in simulation and on the robot.

\textbf{Policy training and execution.}
Both A2A-Policy variants use the same frozen \modelB{} checkpoint across simulation and reality. The three DINOv2~\citep{oquab2023dinov2} variants use $448\times448$ RGB inputs, with region overlays for explicit conditioning. Policy overlays use threshold 0.7, separately from the threshold used to evaluate UAD grounding. Implicit conditioning pools thresholded top-1 \modelB{} features at $336\times336$ in simulation and $448\times448$ on the robot. All four policies share a four-layer multimodal Transformer fusion encoder (width 1024, eight attention heads, feed-forward width 2048) and a depth-six, width-768 diffusion head~\citep{chi2025diffusion,zhao2023learning}. They use two-frame history, seven-dimensional actions, normalized 16-step action chunks, and 100 training diffusion steps. For LIBERO, all variants train for 30 epochs with AdamW (learning rate $10^{-4}$) and per-GPU batch size 16. Policies replan every eight steps using eight DDIM steps. On the Piper arm, each baseline learns one policy across the four tasks, using head/wrist RGB and frozen MiniLM-L6-v2 language features.

App.~\ref{app:policy-implementation} gives the full policy architecture, optimization schedule, and robot deployment interface.
\subsection{A2A-AffordGen: Annotation Quality and Efficiency}
\label{sec:a2a-affordgen-eval}

\textbf{Single-instance grounding.}
Tab.~\ref{tab:affordance-baselines}(a) compares text/box prompting, iterative clicking, and mask decoding under the ORPS protocol. \modelA{} achieves 81.91 gIoU, 80.55 cIoU, 90.50 P@50, and 75.00 P@50--95, leading all four columns. Its gIoU exceeds finetuned LISA-7B by 13.66 points. The improvement also persists at stricter overlap thresholds: P@50--95 exceeds finetuned LISA by 18.97 points. Thus, the difference is not restricted to getting a coarse region above the 0.5 threshold.

\modelB{} achieves 72.78 gIoU under the same ORPS protocol, between finetuned LISA and the annotation teacher. The teacher--student gIoU gap is 9.13 points. Their roles differ: \modelA{} prioritizes annotation quality and takes approximately 15.7 seconds per query, whereas \modelB{} supports online grounding. Evaluating the student on ORPS isolates this quality comparison; its intended task-instruction interface is evaluated separately in Sec.~\ref{sec:exp-modelb}.

\begin{table}[htbp]
\centering\small
\renewcommand{\arraystretch}{1.14}
\caption{ORPS grounding on A2A-Bench, with 200 held-out scenes per protocol. Results averaged over three seeds; percentages except Count MAE.}
\label{tab:affordance-baselines}
\label{tab:multi-scene-overlap}
\label{tab:multi-scene-200}
\textbf{(a) Single-instance grounding}\par\smallskip
\begin{adjustbox}{max width=\linewidth}
\begin{tabular}{@{}lcccc@{}}
\toprule
Method & gIoU & cIoU & P@50 & P@50--95 \\
\midrule
SAM3+text~\citep{carion2025sam} & 44.58 & 46.07 & 48.00 & 38.70 \\
GroundingDINO+SAM3~\citep{liu2024grounding} & 41.87 & 30.27 & 41.17 & 29.65 \\
Qwen3.5-9B+SAM3~\citep{team2026qwen3} & 57.04 & 71.97 & 61.83 & 46.83 \\
Qwen3-VL-32B+SAM3~\citep{bai2025qwen3vl} & 59.25 & 63.33 & 66.83 & 50.73 \\
SegAgent~\citep{zhu2025segagent,bai2023qwen,kirillov2023segment} & 21.62 & 19.51 & 17.83 & 10.10 \\
LISA-7B~\citep{lai2024lisa} & 25.83 & 31.27 & 24.67 & 13.50 \\
LISA-7B (finetuned)~\citep{lai2024lisa} & 68.25 & 76.44 & 74.67 & 56.03 \\
\modelB{} (student) & 72.78 & 78.66 & 76.67 & 57.92 \\
\textbf{\modelA{} (teacher)} & \textbf{81.91} & \textbf{80.55} & \textbf{90.50} & \textbf{75.00} \\
\bottomrule
\end{tabular}
\end{adjustbox}
\par\medskip
\textbf{(b) Multi-scene grounding}\par\smallskip
\begingroup
\footnotesize\setlength{\tabcolsep}{2.5pt}
\begin{adjustbox}{max width=\linewidth}
\begin{tabular}{@{}lcccccccc@{}}
\toprule
Method & gIoU & cIoU & P@50 & P@50--95 & sIoU & Set-R & Set-P & \shortstack{Count\\MAE $\downarrow$} \\
\midrule
SAM3+text~\citep{carion2025sam} & 60.84 & 65.98 & 62.50 & 57.60 & 53.56 & 53.33 & 65.56 & 1.73 \\
GroundingDINO+SAM3~\citep{liu2024grounding} & 40.25 & 29.14 & 32.83 & 23.88 & 30.33 & 21.67 & 33.15 & 1.98 \\
Qwen3.5-9B+SAM3~\citep{team2026qwen3} & 24.48 & 32.14 & 23.17 & 15.10 & 21.75 & 24.32 & 47.13 & 2.60 \\
Qwen3-VL-32B+SAM3~\citep{bai2025qwen3vl} & 38.99 & 59.78 & 41.17 & 24.75 & 32.45 & 33.17 & 55.10 & 2.10 \\
SegAgent~\citep{zhu2025segagent,bai2023qwen,kirillov2023segment} & 29.23 & 23.84 & 16.83 & 11.40 & 21.51 & 12.15 & 24.47 & 2.03 \\
LISA-7B~\citep{lai2024lisa} & 33.32 & 31.60 & 21.83 & 17.55 & 22.82 & 21.97 & 21.39 & 2.46 \\
LISA-7B (finetuned)~\citep{lai2024lisa} & 53.18 & 61.40 & 57.50 & 43.92 & 40.74 & 43.23 & 34.96 & 2.29 \\
\modelB{} (student) & 63.44 & 68.30 & 66.50 & 54.57 & 56.77 & 56.32 & 62.67 & 1.60 \\
\textbf{\modelA{} (teacher)} & \textbf{71.25} & \textbf{73.82} & \textbf{79.50} & \textbf{67.15} & \textbf{67.85} & \textbf{73.88} & \textbf{71.56} & \textbf{1.23} \\
\bottomrule
\end{tabular}
\end{adjustbox}
\endgroup
\end{table}

\textbf{Multi-instance grounding.}
Tab.~\ref{tab:affordance-baselines}(b) reports all baselines on the expanded 200-scene ORPS evaluation. Its overlap metrics measure mask agreement, while the set metrics additionally test coverage, proposal precision, and instance counts. \modelA{} reaches 71.25 gIoU and 73.82 cIoU, together with 67.85 sIoU, 73.88\% set recall, and 71.56\% set precision. Its count MAE is 1.23, the lowest among the listed methods.

Relative to \modelB{}, the teacher gains 11.08 sIoU points and 17.56 set-recall points, while reducing count MAE from 1.60 to 1.23. Relative to native SAM3 text prompting, its sIoU improves by 14.29 points and set recall by 20.55 points. The student also improves overlap and recall over native SAM3, but its set precision of 62.67\% is below SAM3's 65.56\%. This distinction matters: improved coverage can coexist with additional spurious proposals. The full pipeline improves both recall and precision over SAM3, supporting the use of instance refinement and mask-out for annotation. Finetuned LISA improves markedly over its untuned counterpart, but remains behind the teacher on the multi-instance overlap and set metrics.

\textbf{Component ablations.}
Tab.~\ref{tab:ag-ablation} removes one component at a time from the full multi-instance pipeline. Replacing SAM3 text grounding with GroundingDINO boxes causes the largest drop, 24.5 gIoU points, indicating that the initial mask proposals are an important part of the pipeline. Removing iterative clicking costs 10.4 points. Removing direct acceptance costs 4.7 points, while removing the SAG-refine seed mask costs 4.3 points. These results support preserving reliable initial predictions and using uncertain masks as refinement seeds, rather than treating every candidate as a fresh segmentation problem.

\begin{table}[htbp]
\centering\small
\renewcommand{\arraystretch}{1.12}
\caption{Component ablations of A2A-AffordGen on multi-scene ORPS grounding. Each row changes one component of the full pipeline (71.25 gIoU); the reported decrease is in percentage points. Negative values denote lower accuracy.}
\label{tab:ag-ablation}
\begin{adjustbox}{max width=\linewidth}
\begin{tabular}{@{}lc@{}}
\toprule
Change to the full pipeline & $\Delta$gIoU \\
\midrule
Replace SAM3 text grounding with GroundingDINO boxes & $-24.5$ \\
Remove iterative clicking & $-10.4$ \\
Remove direct acceptance & $-4.7$ \\
Remove the SAG-refine seed mask & $-4.3$ \\
Remove dilated gray mask-out & $-2.5$ \\
Reduce the click cap from eight to two & $-3.2$ \\
\bottomrule
\end{tabular}
\end{adjustbox}
\end{table}

Removing dilated gray mask-out costs 2.5 points. This is consistent with the role of local context control: nearby objects can distract a part-level click policy, while dilation preserves a narrow context band around the current region. Reducing the maximum click count from eight to two costs 3.2 points, indicating that some difficult regions require repeated correction. The ablations measure changes in the complete pipeline; they do not imply that the individual drops add linearly.

\textbf{Human verification and correction.}
A separate audit of 500 SAG and 500 MAG masks measures annotation usability. Reviewers accept 76.6\% and 65.4\% directly, with 4.5 and 10.4 seconds of review per mask, respectively. App.~\ref{app:annotation-audit} reports all decision rates, MAG branch proportions, and the verification interface. It distinguishes automatic refinement from human correction, and audit recall from the IoU-thresholded set metric.

Training curves and their interpretation are provided in App.~\ref{app:training-behavior}.
\FloatBarrier
\subsection{A2A-GroundingModel and Zero-shot Transfer}
\label{sec:exp-modelb}

\textbf{Task-conditioned grounding and latency.}
\modelB{} directly maps an RGB image and a TRPS instruction to functional-region masks. Tab.~\ref{tab:modelb-realtime} compares it with A2A-Bench/TRPS-trained UAD~\citep{tang2025uad}, SAM3-I~\citep{li2025sam3}, and SAM3-LoRA~\citep{zhong2024convolution}. Unlike ORPS, the query names an intended action rather than explicitly naming the target part. This evaluation therefore tests whether the model identifies the relevant functional region from manipulation intent.

\textbf{UAD adaptation.}
UAD retains its heatmap architecture and is trained using merged binary masks with BCE+Dice supervision and no target blur. For each query, the union of positive instance masks is resized to the output grid. Dice has weight 1.0 and smoothing $\epsilon=1.0$; $448\times448$ inputs yield $32\times32$ heatmaps. Training uses Adam for 50 epochs, batch size 128, initial learning rate $2\times10^{-3}$, and a decay factor of 0.75 every ten epochs. Grounding evaluation binarizes the heatmap at 0.1. This threshold belongs to segmentation evaluation and differs from the 0.7 overlay threshold used by the downstream policies. App.~\ref{app:uad-baseline} gives the loss definitions, training settings, and original UAD diagnostic experiments; those earlier diagnostics are reported separately from the current benchmark.

\begin{table}[htbp]
\centering\small
\renewcommand{\arraystretch}{1.12}
\caption{Task-conditioned TRPS grounding on 200 fixed mask--query pairs, averaged over three seeds. Accuracy scores are percentages. Latency and throughput are measured at $640\times480$ input resolution; lower latency and higher FPS are better.}
\label{tab:modelb-realtime}
\begin{adjustbox}{max width=\linewidth}
\begin{tabular}{@{}lcccccc@{}}
\toprule
Method & gIoU & cIoU & P@50 & P@50--95 & ms $\downarrow$ & FPS $\uparrow$ \\
\midrule
UAD~\citep{tang2025uad} & 28.45 & 36.39 & 28.17 & 10.03 & \textbf{62} & \textbf{16.1} \\
SAM3-I~\citep{li2025sam3} & 50.99 & 47.03 & 51.00 & 35.10 & 101 & 9.9 \\
SAM3-LoRA~\citep{zhong2024convolution} & 52.98 & 46.26 & 53.67 & 36.97 & 105 & 9.5 \\
\textbf{\modelB{}} & \textbf{55.41} & \textbf{49.52} & \textbf{56.83} & \textbf{39.55} & 103 & 9.7 \\
\bottomrule
\end{tabular}
\end{adjustbox}
\end{table}

\modelB{} leads the four reported accuracy metrics, reaching 55.41 gIoU and 49.52 cIoU at 103 ms per image (9.7 FPS). Against SAM3-LoRA, it improves gIoU by 2.43 points and cIoU by 3.26 points with a similar reported latency. UAD is faster at 62 ms (16.1 FPS), but its gIoU is 28.45. These results expose the accuracy--latency tradeoff rather than identifying one method as best on every axis. The reported inference rate measures grounding, not complete closed-loop robot execution.

\begin{figure}[htbp]
\centering
\includegraphics[width=\linewidth]{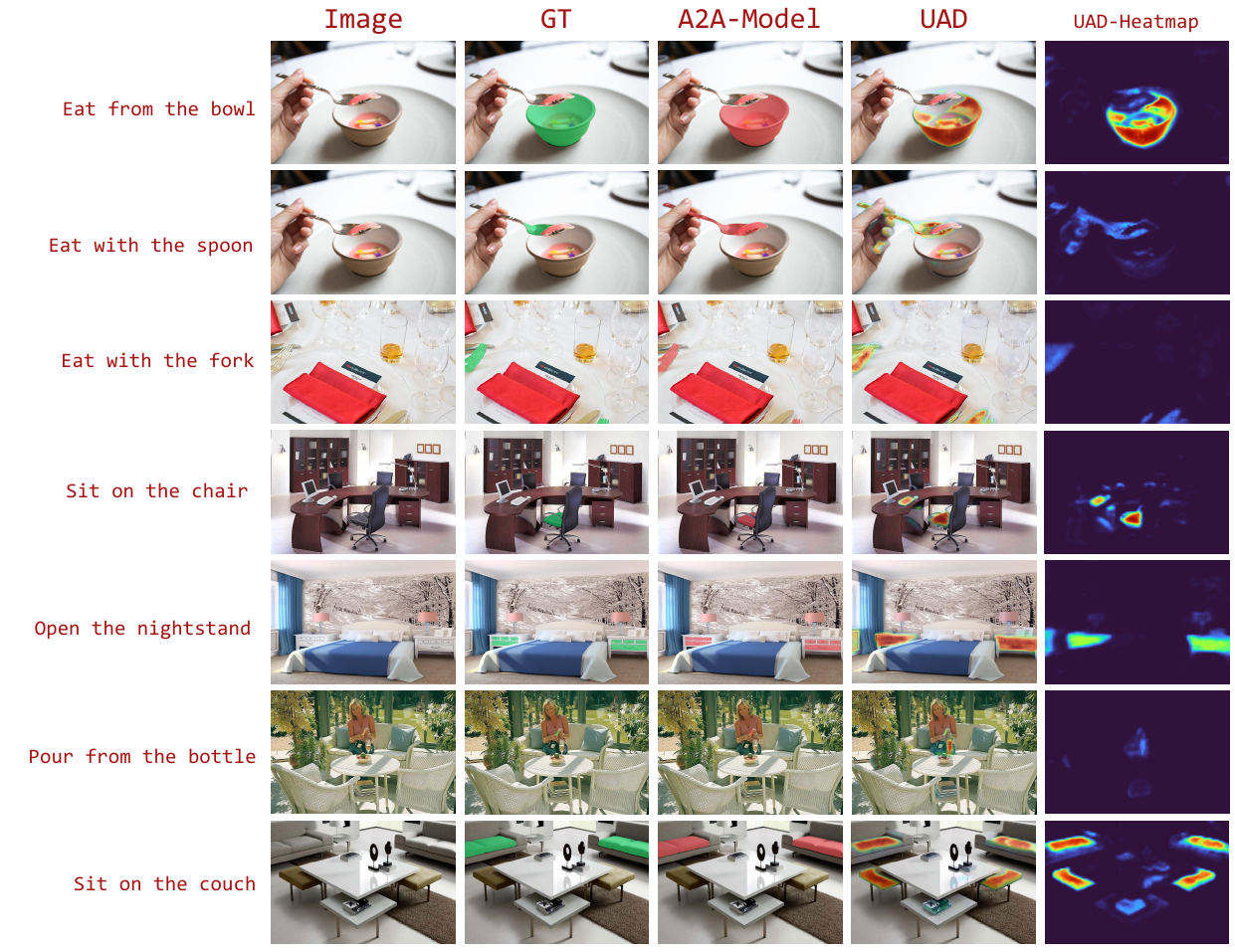}
\caption{Task-conditioned masks on A2A-Bench. The seven examples show the input image and task, ground-truth mask, \modelB{} prediction (labeled A2A-Model), binarized UAD prediction, and UAD heatmap. Tasks include eating, sitting, opening, and pouring. The heatmaps expose diffuse responses on unrelated regions, complementing the quantitative TRPS evaluation.}
\label{fig:a2a-model-qualitative}
\end{figure}

Fig.~\ref{fig:a2a-model-qualitative} illustrates the distinction between diffuse heatmap responses and task-aligned part masks. Small handles, seats, and object contact regions require localization within an object, rather than merely identifying its category. The examples visualize typical spatial differences, while aggregate performance is measured on the held-out set.

\begin{figure}[htbp]
\centering
\includegraphics[width=0.94\linewidth,height=0.70\textheight,keepaspectratio]{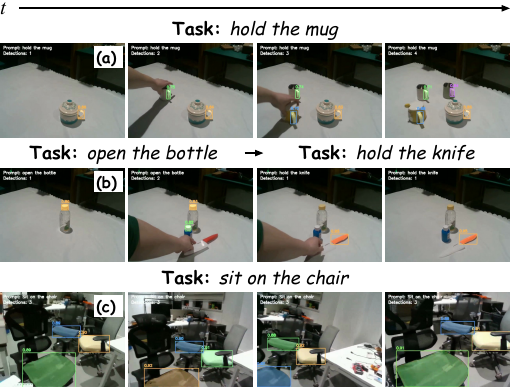}
\caption{Live A2A-GroundingModel demonstrations. (a) Additional mugs introduce new target handles. (b) The instruction changes from opening bottles to holding a knife. (c) Camera motion changes the chair views. Frames progress from left to right within each sequence.}
\label{fig:live-grounding-demos}
\end{figure}

\textbf{Live real-world grounding.}
Fig.~\ref{fig:live-grounding-demos} shows predictions as scene content, the instruction, and the camera view change. The first sequence adds mugs while the instruction remains fixed, testing whether the model identifies multiple relevant handles. The second changes the target from bottle caps to a knife handle through an instruction change. The third changes the viewpoint around chairs while targeting their seats. These demonstrations illustrate live responses to changing observations; they are qualitative sequences, not a separate success-rate benchmark.

\textbf{Zero-shot referring segmentation.}
We evaluate unchanged \modelA{} and \modelB{} checkpoints on RefCOCO, RefCOCO+, and RefCOCOg without fine-tuning on these datasets. Tab.~\ref{tab:zero-shot-main} separates the zero-shot methods from models trained with benchmark-specific supervision. The fine-tuned group contextualizes performance but does not share the same training regime.

\begin{table}[htbp]
\centering\small
\renewcommand{\arraystretch}{1.12}
\caption{Referring segmentation (cIoU, \%). Fine-tuned methods use benchmark-specific training and provide context; the zero-shot group uses unchanged checkpoints without RefCOCO-family fine-tuning. The two groups represent different supervision regimes.}
\label{tab:zero-shot-main}
\begin{tabular*}{\columnwidth}{@{\extracolsep{\fill}}lccc@{}}
\toprule
Method & RefCOCO & RefCOCO+ & RefCOCOg \\
\midrule
\multicolumn{4}{@{}l}{\emph{Fine-tuned}} \\
LAVT~\citep{yang2022lavt}                   & 72.7  & 62.1  & 61.2  \\
CRIS~\citep{wang2022cris}                   & 70.5  & 65.3  & 59.9  \\
PolyFormer-L~\citep{liu2023polyformer}      & 76.94 & \textbf{72.15} & 71.15 \\
SEEM~\citep{zou2023segment}                    & --    & --    & 65.7  \\
LISA(SAM)~\citep{lai2024lisa}              & 74.9  & 65.1  & 67.9  \\
PixelLM~\citep{ren2024pixellm}              & 73.0  & 66.3  & 69.3  \\
PerceptionGPT~\citep{pi2024perceptiongpt}   & 75.1  & 68.5  & 70.3  \\
GSVA(SAM)~\citep{xia2024gsva}              & 77.2  & 65.9  & 72.7  \\
SAM4MLLM(Qwen)~\citep{chen2024sam4mllm}    & 77.1  & 71.5  & \textbf{74.5}  \\
SegAgent-Qwen+SAM~\citep{zhu2025segagent}         & \textbf{78.01} & 70.86 & 74.49 \\
\midrule
\multicolumn{4}{@{}l}{\emph{Zero-shot}} \\
SAM3+text~\citep{carion2025sam} & 39.40 & 29.44 & 32.76 \\
\modelB{} & 47.34 & 37.28 & 48.66 \\
\textbf{\modelA{}} & \textbf{75.86} & \textbf{64.78} & \textbf{72.55} \\
\bottomrule
\end{tabular*}
\end{table}

\modelA{} reaches 75.86, 64.78, and 72.55 cIoU on the three benchmarks, exceeding text-prompted SAM3 by 36.46, 35.34, and 39.79 points. \modelB{} reaches 47.34, 37.28, and 48.66, improving over the same baseline by 7.94, 7.84, and 15.90 points. The student retains positive transfer relative to its underlying text-prompted baseline, although a substantial teacher--student gap remains.

Among the listed fine-tuned models, \modelA{} exceeds LAVT, LISA, PixelLM, and PerceptionGPT on RefCOCO and RefCOCOg. On RefCOCO+, it is 0.32 points below LISA and below several other fine-tuned baselines. Thus, the transfer is useful across all three datasets without implying uniform superiority over benchmark-specific training. Moving from functional-part supervision to whole-object referring expressions tests a different target granularity, and the results suggest that the learned click procedure transfers beyond the part phrases used to construct A2A-Bench.
\FloatBarrier
\subsection{A2A-Policy: Simulation and Real-robot Manipulation}
\label{sec:exp-a2a-policy}

\textbf{Controlled policy comparison.}
All four policies share the Diffusion Policy head~\citep{chi2025diffusion} and Transformer fusion of visual, language, and state tokens. DP-RGB uses frozen DINOv2 without affordances. UAD-DP overlays UAD heatmaps on the RGB observation. A2A-Explicit overlays \modelB{} masks before RGB feature extraction. A2A-Implicit instead uses \modelB{} features with Afford-VLA-style~\citep{wang2026afford} top-1 selection, thresholding, and mask pooling. Both A2A variants use the same frozen \modelB{} in simulation and on the robot; \modelA{} does not enter the policy loop.

The implicit interface replaces unfiltered coarse FPN pooling while keeping the training setup unchanged. It restricts spatial aggregation to a predicted task-relevant region instead of asking the policy to select useful features from the entire coarse feature map. The comparison therefore tests two interfaces to the same grounding prior: explicit observation highlighting and implicit region-conditioned feature pooling.

\begin{table}[htbp]
\centering\small
\renewcommand{\arraystretch}{1.12}
\caption{Policy evaluation with matched action-head and fusion architectures. LIBERO-Object reports the mean success rate over ten tasks; the real robot reports successes across four tasks with ten trials each. Higher is better.}
\label{tab:policy-eval}
\begin{adjustbox}{max width=\linewidth}
\begin{tabular}{@{}lcc@{}}
\toprule
Policy & LIBERO-Object (\%) & Piper arm (successes/trials) \\
\midrule
DP-RGB~\citep{chi2025diffusion} & 87.8 & 22/40 \\
UAD-DP~\citep{tang2025uad} & 74.4 & 16/40 \\
\textbf{A2A-Explicit} & \textbf{94.4} & \textbf{26/40} \\
A2A-Implicit & 92.8 & 25/40 \\
\bottomrule
\end{tabular}
\end{adjustbox}
\end{table}

\textbf{LIBERO-Object aggregate results.}
A2A-Explicit and A2A-Implicit achieve 94.4\% and 92.8\% mean success, compared with 87.8\% for DP-RGB and 74.4\% for UAD-DP (Tab.~\ref{tab:policy-eval}). Their absolute improvements over DP-RGB are 6.6 and 5.0 percentage points. The corresponding reductions in residual failure rate are
\begin{equation}
\frac{94.4-87.8}{100-87.8}\approx54.1\%,\qquad
\frac{92.8-87.8}{100-87.8}\approx41.0\%.
\label{eq:policy-failure-reduction}
\end{equation}
These are derived from the displayed aggregate success rates, not averages of per-task or per-seed relative reductions. Both interfaces improve on the RGB baseline, while explicit highlighting retains the strongest aggregate performance.

\textbf{Object-wise variation.}
Explicit conditioning improves cream cheese from 90\% to 98\%, barbecue sauce from 72\% to 86\%, and chocolate pudding from 88\% to 100\%; milk remains at 92\%. App.~\ref{app:policy-breakdowns} provides the available per-task scores, whose means match Tab.~\ref{tab:policy-eval}. Only the aggregate is available for the updated implicit policy.

\textbf{Real-robot results.}
A2A-Explicit achieves 26/40 successful trials and A2A-Implicit 25/40, compared with 22/40 for DP-RGB and 16/40 for UAD-DP. These totals correspond to 65.0\%, 62.5\%, 55.0\%, and 40.0\%, respectively. Explicit highlighting adds four successes over DP-RGB, while implicit pooling adds three. The per-task counts for the RGB and explicit-conditioning policies appear in App.~\ref{app:policy-breakdowns}.

Explicit highlighting succeeds in nine microwave-opening, eight stacking, seven cup-placement, and two phone-placement trials, out of ten for each task. The corresponding DP-RGB counts are eight, six, six, and two. Phone placement remains difficult, with no improvement in these observed counts. A better functional-region prior can improve target selection without resolving contact, pose alignment, and action-execution difficulty. With ten trials per task, the individual counts are descriptive outcomes of this evaluation; we do not infer statistical significance from these totals alone.

\textbf{Qualitative execution sequences.}
Figs.~\ref{fig:realworld-demo-stack}--\ref{fig:realworld-demo-cup} in App.~\ref{app:robot-demos} show the four real-world demonstrations. Each figure follows the same row order, placing UAD-DP above A2A-Implicit and A2A-Explicit, with time advancing from left to right. The comparisons reveal how an affordance signal enters the policy observation or representation during execution. These selected sequences illustrate behavior and do not replace the complete trial counts.

Real-world distribution shifts can weaken UAD heatmaps or highlight an incorrect part, making an overlay misleading for the action model. Task-aligned masks provide a more targeted observation prior, while thresholded region pooling makes the implicit interface competitive without an overlay. Nevertheless, the unchanged phone-placement count and remaining failures in the other tasks show that grounding accuracy alone does not guarantee successful manipulation.

\section{Conclusion and Limitation}

\textbf{Affordance2Action (A2A)} connects scalable affordance annotation, real-time scene-level grounding, and robotic manipulation. \textbf{A2A-AffordGen} reduces annotation effort through high-quality masks and lightweight verification; \textbf{A2A-GroundingModel} improves task-conditioned localization at $9.7$ FPS. With frozen grounding, \textbf{A2A-Policy}'s explicit variant raises success from $87.8\%$ to $94.4\%$ on LIBERO-Object and from $22/40$ to $26/40$ on the real robot. Its implicit variant reaches $92.8\%$ and $25/40$ with thresholded region pooling, showing that both tested interfaces benefit from the frozen grounding prior. The 200-scene multi-instance evaluation, component ablations, and human-verification audit jointly characterize annotation accuracy, instance coverage, and review effort.
Despite these gains, broader deployment requires addressing the following limitations. Policy evaluation covers simulation and real Piper-arm tabletop tasks. Mobile and whole-body manipulation, including navigation, base placement, and coordinated arm execution in larger scenes, remain untested. Robot-induced occlusion, object motion, and contact disturbances can reduce affordance reliability. Beyond the current threshold-based filtering, dynamic grounding for closed-loop interaction remains future work.

\FloatBarrier
\clearpage
\bibliographystyle{unsrtnat}
\bibliography{references}

\clearpage
\appendix
\section{A2A-AffordGen: Training and Inference Details}
\label{app:affordgen-details}

This section expands the construction and inference procedures in Sec.~\ref{sec:a2a_bench}. The learned component is the click policy; SAM3 supplies the mask transition, and MAG supplies the scene-level wrapper. These components have different responsibilities and do not share a single end-to-end training loss.

\subsection{Supervision and Serialized Click Targets}

The training corpus combines approximately 40K curated object-part masks with approximately 18K manually annotated part masks from the 5,000-scene core, giving approximately 58K object-part masks before trajectory expansion. The oracle converts these masks into approximately 150K click-trajectory samples. The 5,000 automatically generated, human-verified scenes form the initial released batch; they are distinct from the manually annotated core and from the held-out evaluation scenes. A scene may contribute several functional regions and therefore several crop-level training examples.

For a crop $I$, explicit part prompt $P$, and current mask $M_t$, the policy state contains the image with its current-mask overlay and the prompt. The reference mask $M^\star$ is used to produce supervision. The two residual sets,
\begin{equation}
E_t^+=M^\star\setminus M_t,\qquad E_t^-=M_t\setminus M^\star,
\end{equation}
separate missed target pixels from incorrectly included pixels. The oracle chooses the larger residual set and places a click at its distance-transform maximum, as in Eq.~\ref{eq:sag-oracle}. Positive and negative clicks therefore have distinct roles even when they occur on the same object. After each oracle action, frozen SAM3 materializes the next mask; the next training state reflects that updated mask rather than reusing the initial prediction.

The target sequence contains four fields: click polarity, click location, next-mask IoU, and incremental IoU gain. One example of the structured output is \emph{``Positive point: $(x,y)$. Next IoU: $v$. Gain: $\delta$.''} The quality targets are measured after applying the oracle click:
\begin{equation}
v_t^\star=\mathrm{IoU}(M_{t+1}^\star,M^\star),\qquad
\delta_t^\star=v_t^\star-\mathrm{IoU}(M_t,M^\star),
\end{equation}
where $M_{t+1}^\star=\Phi(I,P,M_t,a_t^\star)$ denotes the result of the oracle action, not a second ground-truth mask. This distinguishes the target quality of the proposed next state from the quality of the current state.

Qwen3.5-9B is adapted with LoRA to predict the serialized sequence $y_t^\star$. Writing its tokens as $y_{t,1}^\star,\ldots,y_{t,L_t}^\star$, the token-level cross-entropy in Eq.~\ref{eq:sag-loss} has the autoregressive form
\begin{equation}
\mathcal L_{\mathrm{agent}}(s_t,y_t^\star)
=-\sum_{j=1}^{L_t}\log p_\theta(y_{t,j}^\star\mid s_t,y_{t,<j}^\star).
\label{eq:app-click-ce}
\end{equation}
Thus, the click and stopping estimates are learned through the same structured text target. No pixel-level segmentation loss is back-propagated through SAM3 into the language model. Oracle training rollouts terminate at the reference-IoU target $\kappa$ or the click budget. At deployment, the reference mask and oracle residual sets are unavailable.

\subsection{Stopping Order and Seeded Refinement}

SAG can start from an empty mask or from a low-confidence SAM3 mask. In both cases, the same trained policy receives the current mask and proposes a correction. Alg.~\ref{alg:affordgen-inference} uses an in-place mask update and permits at most eight clicks. The order matters because the predicted quality describes the next mask:
\begin{enumerate}[leftmargin=1.5em,itemsep=3pt,topsep=3pt]
\item Query the policy for $(p_t,u_t,\hat v_t,\hat\delta_t)$ using the current mask.
\item If $\hat\delta_t<\Delta_{\min}$, stop and return the current mask without applying the proposed click.
\item Otherwise, apply the click through frozen SAM3 and replace the current mask with the result.
\item If $\hat v_t>\tau_{\mathrm{iou}}$, stop with that updated mask; otherwise continue until the click budget is exhausted.
\end{enumerate}
The reference-IoU target used to construct training rollouts is separate from these predicted stopping signals. A seed mask also changes the starting state, not the objective or policy parameters. Removing the seed in the ablation forces refinement to reconstruct the region from an empty mask and costs 4.3 gIoU points in the full pipeline.

\subsection{MAG Crops, Mask-out, and Scene Reassembly}

SAM3 text prompting returns candidate masks and confidence scores, while Grounding DINO supplies object boxes. A confident candidate bypasses SAG. A low-confidence candidate supplies both an initial mask and a matched crop for seeded refinement. An unmatched object box starts SAG from an empty mask, including when SAM3 has returned no mask for the scene. The matching and crop wrapper allow the single-instance policy to be reused without training a separate multi-instance click model.

For seeded refinement, let $D_i=m_i\oplus\mathcal K$ be the dilated candidate region and let $g$ denote the neutral gray background. The image presented to the crop operator is
\begin{equation}
I_i^{\mathrm{gray}}(u)=
\begin{cases}
I(u), & u\in D_i,\\
g, & u\notin D_i.
\end{cases}
\qquad
I_{b_i}=\mathcal C_{b_i}(I_i^{\mathrm{gray}}).
\label{eq:app-gray-crop}
\end{equation}
The initial mask is transformed with the same crop operator, $M_0=\mathcal C_{b_i}(m_i)$. Dilation preserves a local context band, while gray mask-out suppresses neighboring objects. An unmatched box has no seed mask, so its image is cropped directly from $I$ and its rollout starts from $M_0=\emptyset$.

Each refined crop mask is mapped back with $\mathcal C_{b_i}^{-1}$. Directly accepted masks already use scene coordinates and require the identity mapping. Reassembly retains the individual masks as an instance set, as in Eq.~\ref{eq:mag-reassembly}; it does not discard instance identities by collapsing all output regions into one binary target. This distinction matters for the padded set matching and count error in Sec.~\ref{sec:exp-setup}.

Automatic refinement is distinct from the subsequent human verification stage. MAG chooses whether to accept, refine, or initialize an instance; a human reviewer subsequently decides whether the resulting annotation can be retained, corrected, or rejected. App.~\ref{app:annotation-audit} reports both processes with their respective denominators.

\subsection{Training Behavior}
\label{app:training-behavior}
\begin{figure}[htbp]
\centering
\includegraphics[width=0.90\linewidth]{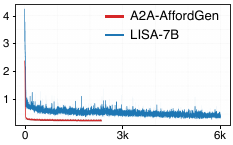}
\caption{Training trends of A2A-AffordGen and LISA-7B. The horizontal axis is the training step; the vertical axis shows each model's own training objective. Red triangles mark A2A-AffordGen validation probes. Different objectives prevent interpreting the absolute loss gap as a quality comparison.}
\label{fig:training-curves}
\end{figure}

Fig.~\ref{fig:training-curves} shows the training traces with matched A2A-Bench splits and optimizer settings. \modelA{} learns a serialized click-policy target against frozen SAM3, whereas LISA-7B~\citep{lai2024lisa} optimizes its own segmentation objective. The curves therefore visualize optimization within each model rather than a common loss scale.

\modelA{} shows a rapid initial decrease followed by a sustained plateau with limited fluctuations. LISA also decreases initially, then improves more gradually with step-to-step variation. These traces describe optimization behavior for the displayed runs. Grounding quality is established by the held-out mask and set metrics in Tab.~\ref{tab:affordance-baselines}, rather than by comparing absolute loss values or inferring cross-seed stability from one plot.

\FloatBarrier
\section{A2A-GroundingModel: Detailed Formulation}
\label{app:grounding_model_details}

This section provides the full formulation of A2A-GroundingModel, including staged instruction adaptation, text-conditioned visual prompt injection, and the training objective. The main text gives the core equations; the step-by-step formulation below separates the query routes, prompt tensors, and training targets.

\subsection{Staged Instruction Adaptation}

A2A-GroundingModel uses two types of language supervision. Oracle queries explicitly describe the target functional part, while task queries describe the manipulation intent and require the model to infer the corresponding actionable region. To reduce the interference between explicit part recognition and task-level reasoning, we introduce separate but connected adaptation paths for oracle and task instructions. Routing is fixed by the supplied query type rather than selected by a learned gate. ORPS uses the O-Adapter route; TRPS uses the oracle-to-task bridge and T-Adapter route. Only the task route is used at deployment.

Let $\mathbf{x}^{(\ell)}$ denote the hidden state at the $\ell$-th adapterized block. For oracle part descriptions, we use a lightweight bottleneck adapter $A_o^{(\ell)}$:
\begin{equation}
\mathbf{x}^{(\ell)}_o =
\mathbf{x}^{(\ell)}
+
A_o^{(\ell)}
\left(
\mathrm{LN}(\mathbf{x}^{(\ell)})
\right).
\end{equation}

\begin{samepage}
For task instructions, we introduce an oracle-to-task adaptation path. The hidden state is first transformed by an oracle-to-task bridge adapter $A_{o\rightarrow t}^{(\ell)}$, and then further refined by a task-specific adapter $A_t^{(\ell)}$:
\begin{equation}
\begin{aligned}
\mathbf r^{(\ell)}&=\mathbf x^{(\ell)}+A_{o\to t}^{(\ell)}\!\left(\mathrm{LN}(\mathbf x^{(\ell)})\right),\\
\mathbf x_t^{(\ell)}&=\mathbf r^{(\ell)}+A_t^{(\ell)}\!\left(\mathrm{LN}(\mathbf r^{(\ell)})\right).
\end{aligned}
\end{equation}
\end{samepage}

This staged design allows the model to first acquire stable part-level grounding from explicit part queries and then transfer this knowledge to task-level affordance reasoning. We apply the same adaptation strategy to selected language, fusion, decoder, and mask prediction blocks.

\subsection{Text-Conditioned Visual Prompt Injection}

Staged instruction adaptation improves task understanding, but task-conditioned affordance grounding also requires accurate localization of small and visually subtle functional parts. To inject task information into the visual representation, we introduce a text-conditioned visual prompt generator.

Given token-level language features
$\mathbf{T}\in\mathbb{R}^{L\times d_t}$, we obtain a task representation
$\mathbf{z}\in\mathbb{R}^{d_t}$
by masked average pooling over valid text tokens. With $\mathcal I_q$ denoting the valid, non-padding token positions,
\begin{equation}
\mathbf z=\frac{1}{|\mathcal I_q|}\sum_{i\in\mathcal I_q}\mathbf T_i.
\end{equation}
In parallel, the image is patch-projected and processed by a shallow visual projection network to produce spatial prompt tokens
$\mathbf{V}\in\mathbb{R}^{HW\times d_p}$.
The task representation is mapped to a text bias and a channel-wise gate:
\begin{equation}
\mathbf{b}
=
W_b\,\mathrm{LN}(\mathbf{z}),
\qquad
\mathbf{g}
=
\sigma
\left(
W_g\,\mathrm{LN}(\mathbf{z})
\right),
\end{equation}
where $\mathbf{b},\mathbf{g}\in\mathbb{R}^{d_p}$.

We then compute a text-guided spatial weighting over visual prompt tokens:
\begin{equation}
\alpha_j =
\frac{
\exp
\left(
\mathbf{V}_j^\top \mathbf{b}/\sqrt{d_p}
\right)
}{
\sum_k
\exp
\left(
\mathbf{V}_k^\top \mathbf{b}/\sqrt{d_p}
\right)
},
\end{equation}
where $j$ indexes the spatial location.

The resulting text-conditioned prompt seed is defined as
\begin{equation}
\mathbf{S}_j =
\mathrm{LN}
\left(
\mathbf{V}_j \odot (1+\alpha_j\mathbf{g}) + \mathbf{b}
\right).
\end{equation}

Layer-specific projections then transform $\mathbf{S}$ into prompt tensors
$\mathbf{P}^{(\ell)}$
that match the channel dimension of selected visual encoder blocks. For each prompted layer $\ell$, the visual tokens are updated through additive prompt injection:
\begin{equation}
\mathbf{X}^{(\ell)}
\leftarrow
\mathbf{X}^{(\ell)}
+
\gamma_\ell
\mathbf{P}^{(\ell)},
\end{equation}
where $\gamma_\ell$ is a learnable scaling coefficient. We inject prompts into later visual encoder blocks, where visual features are more semantically structured and therefore more suitable for task-conditioned modulation.

\subsection{Training Objective}

During training, the base model is kept frozen and only the newly introduced adaptation modules are optimized. The model is trained with both oracle part queries and task-level instructions from A2A-Bench. Oracle queries provide explicit part-level supervision, while task queries require grounding the functional region from manipulation intent.

Predicted masks are matched to ground-truth masks by Hungarian assignment. The base segmentation objective is denoted as
$\mathcal{L}_{\mathrm{seg}}$,
which includes classification, box regression, generalized-IoU, and mask supervision terms:
\begin{equation}
\mathcal{L}_{\mathrm{seg}}
=
\lambda_{\mathrm{cls}}\mathcal{L}_{\mathrm{cls}}
+
\lambda_{\mathrm{box}}\mathcal{L}_{\mathrm{box}}
+
\lambda_{\mathrm{giou}}\mathcal{L}_{\mathrm{giou}}
+
\lambda_{\mathrm{mask}}\mathcal{L}_{\mathrm{mask}}.
\end{equation}

To align explicit part grounding and task-level affordance grounding, we further impose oracle--task consistency regularization. Let
$\mathbf{P}_o$
and
$\mathbf{P}_t$
denote the spatial affordance distributions predicted from the oracle query and task query, respectively. We use a bidirectional KL divergence:
\begin{equation}
\mathcal{L}_{\mathrm{KL}}
=
\frac{1}{2}
\left[
D_{\mathrm{KL}}(\mathbf{P}_o \Vert \mathbf{P}_t)
+
D_{\mathrm{KL}}(\mathbf{P}_t \Vert \mathbf{P}_o)
\right].
\end{equation}

We also introduce a disagreement-weighted task-query loss to emphasize pixels where the oracle and task predictions are inconsistent. Let
$\mathbf{W}_{\mathrm{task}}$
be a disagreement weight map computed from the prediction difference between the two routes. The task-query loss is written as
\begin{equation}
\mathcal{L}_{\mathrm{task}}
=
\frac{1}{|\Omega|}
\sum_{u\in\Omega}
\mathbf{W}_{\mathrm{task}}(u)
\cdot
\mathcal{L}_{\mathrm{mask}}
\left(
\hat{M}_{\mathrm{task}}(u), M(u)
\right),
\end{equation}
where $\Omega$ denotes the image lattice, $\hat{M}_{\mathrm{task}}$ is the task-query prediction, and $M$ is the ground-truth affordance mask.

The final training objective is
\begin{equation}
\mathcal{L}
=
\mathcal{L}_{\mathrm{seg}}
+
\lambda_{\mathrm{KL}}
\mathcal{L}_{\mathrm{KL}}
+
\lambda_{\mathrm{task}}
\mathcal{L}_{\mathrm{task}}.
\end{equation}

This objective encourages the model to preserve accurate part-level localization from explicit oracle queries while producing spatially consistent predictions under task-level instructions.

\textbf{Training schedule.}
We follow a three-stage oracle-to-task curriculum.
In Stage I, the model is trained with oracle part queries to learn explicit part-level grounding.
In Stage II, the task route is trained with task-level instructions to transfer this grounding ability to affordance reasoning.
In Stage III, oracle and task routes are jointly optimized with an additional oracle--task consistency loss to align their spatial predictions.
Throughout training, the base model is frozen and only the inserted adaptation modules are updated.
\section{UAD Adaptation and Original Diagnostic Experiments}
\label{app:uad-baseline}

\subsection{Supervision Interface and Training Loss}

UAD retains its heatmap architecture. To adapt its supervision to A2A-Bench, each image is paired with its task query, stored as \texttt{complex\_query} in the original data interface. All positive instance masks associated with that query are merged by pixel-wise union into one binary target, which is resized to the UAD output grid. This creates a heatmap-supervision target without changing the UAD model into an instance-set predictor.

Let $p_j\in[0,1]$ be the predicted affordance probability and $y_j\in[0,1]$ the target at output location $j$. With $M$ spatial locations, the binary cross-entropy is
\begin{equation}
\mathcal L_{\mathrm{BCE}}=-\frac{1}{M}\sum_{j=1}^{M}
\left[y_j\log p_j+(1-y_j)\log(1-p_j)\right].
\label{eq:app-uad-bce}
\end{equation}
The soft Dice loss is
\begin{equation}
\mathcal L_{\mathrm{Dice}}=1-
\frac{2\sum_jp_jy_j+\epsilon}{\sum_jp_j+\sum_jy_j+\epsilon},
\qquad \epsilon=1.0.
\label{eq:app-uad-dice}
\end{equation}
The original comparison includes a BCE-only variant and a variant trained with
\begin{equation}
\mathcal L_{\mathrm{UAD}}=\mathcal L_{\mathrm{BCE}}+\lambda\mathcal L_{\mathrm{Dice}},
\qquad \lambda=1.0.
\label{eq:app-uad-total}
\end{equation}
For BCE+Dice, target-mask blur is disabled to match the binary annotations. Tab.~\ref{tab:uad_affordance2act_config} lists the training configuration. The output grid is smaller than the input image; target projection and heatmap thresholding therefore form part of the evaluation interface.

\begin{table}[htbp]
    \centering
    \small
    \caption{Training settings for the adapted UAD baseline on A2A-Bench.}
    \label{tab:uad_affordance2act_config}
    \setlength{\tabcolsep}{6pt}
    \begin{tabular}{ll}
        \toprule
        Item & Setting \\
        \midrule
        Input resolution & $448 \times 448$ \\
        Output resolution & $32 \times 32$ \\
        Optimizer & Adam \\
        Learning rate & $2 \times 10^{-3}$ \\
        Batch size & $128$ \\
        Training epochs & $50$ \\
        LR schedule & Step decay every 10 epochs with factor $0.75$ \\
        \bottomrule
    \end{tabular}
\end{table}

\subsection{BCE versus BCE+Dice}

The following two tables preserve the diagnostic experiments reported in the first manuscript. Their evaluation manifest and seed aggregation are not specified in the available record, so these results are reported separately from the current 200-pair, three-seed TRPS benchmark in Tab.~\ref{tab:modelb-realtime}. They explain the original UAD adaptation and threshold choice; they are not additional matched runs on the current benchmark.

Tab.~\ref{tab:uad_affordance2act_results} compares predictions thresholded at 0.5 with the projected target mask. Here 0.5 is the heatmap binarization threshold, not the per-example IoU threshold used by P@50. The BCE and Dice-loss columns evaluate soft predictions; the remaining columns evaluate binary masks. Consequently, one minus the displayed soft Dice loss need not equal the displayed binary Dice score.

\begin{table}[htbp]
    \centering
    \footnotesize
    \setlength{\tabcolsep}{3pt}
    \caption{Original UAD supervision comparison at heatmap threshold 0.5. BCE and Dice loss use soft predictions; binary overlap, precision, and recall are percentages. These diagnostic results are separate from the current 200-pair benchmark.}
    \label{tab:uad_affordance2act_results}
    \begin{adjustbox}{max width=\linewidth}
    \begin{tabular}{lcccccc}
        \toprule
        Variant & BCE $\downarrow$ & Dice Loss $\downarrow$ & IoU $\uparrow$ & Dice $\uparrow$ & Precision $\uparrow$ & Recall $\uparrow$ \\
        \midrule
        UAD-BCE & \textbf{0.1015} & 0.7634 & 13.92 & 18.55 & \textbf{76.55} & 17.14 \\
        UAD-BCE+Dice & 0.1224 & \textbf{0.6749} & \textbf{25.76} & \textbf{33.40} & 60.93 & \textbf{34.99} \\
        \bottomrule
    \end{tabular}
    \end{adjustbox}
\end{table}

BCE-only training has lower BCE loss and higher binary precision, but recovers a smaller fraction of the target pixels. With BCE+Dice, IoU increases from 13.92\% to 25.76\%, Dice from 18.55\% to 33.40\%, and recall from 17.14\% to 34.99\%, while precision decreases from 76.55\% to 60.93\%. This is a coverage--precision tradeoff rather than an improvement in every metric. The BCE objective's lower numerical value does not by itself establish a better binary mask, and the soft-loss columns should not be compared directly with the percentage-valued overlap columns.

\subsection{Heatmap Threshold Sensitivity}
\label{app:uad-threshold-sensitivity}

Tab.~\ref{tab:uad-threshold-sensitivity} reports the original sweep for BCE+Dice without target blur. Among the displayed settings, 0.1 gives the highest gIoU and P@50, 0.2 the highest cIoU, and 0.3 the highest P@50--95. The metric-specific optima show that selecting a threshold for mean mask overlap need not maximize pooled overlap or success at stricter IoU cutoffs. The mean-overlap score falls from 28.45\% at threshold 0.1 to 15.77\% at threshold 0.9, consistent with increasing suppression of heatmap responses.

\begin{table}[htbp]
\centering
\small
\setlength{\tabcolsep}{6pt}
\renewcommand{\arraystretch}{1.08}
\caption{Original UAD threshold diagnostic with BCE+Dice supervision and no target blur. Scores are percentages; the original values are retained separately from the current 200-pair, three-seed results.}
\label{tab:uad-threshold-sensitivity}

\begin{tabular}{ccccc}
\toprule
Threshold & gIoU (\%) & cIoU (\%) & P@50 (\%) & P@50-95 (\%) \\
\midrule
0.1 & \textbf{28.45} & 36.39 & \textbf{28.22} & 10.03 \\
0.2 & 28.23 & \textbf{36.71} & 27.18 & 10.77 \\
0.3 & 27.70 & 36.43 & 26.83 & \textbf{11.01} \\
0.5 & 25.76 & 34.66 & 24.74 & 10.52 \\
0.7 & 22.27 & 31.33 & 18.82 & 8.71 \\
0.9 & 15.77 & 23.96 & 14.63 & 5.44 \\
\bottomrule
\end{tabular}
\end{table}

The current grounding evaluation uses threshold 0.1, while the downstream UAD-DP overlay uses 0.7. These thresholds serve different purposes. The original threshold-0.1 row reports P@50 of 28.22\%; the current 200-pair benchmark reports 28.17\%. We retain each table's own recorded value rather than combining the two evaluations. Likewise, the 25.76\% IoU in the original BCE+Dice comparison agrees with the threshold-0.5 sweep row, providing a cross-check within those original diagnostics.

\section{Human Verification and Annotation Routing}
\label{app:annotation-audit}

\textbf{Human verification and correction.}
We audit 500 newly generated single-instance Objects365 masks and 500 newly generated multi-instance masks. Reviewers use per-part click correction, threshold adjustment, accept/reject decisions, and missed-region counting. These 500-mask audits assess annotation usability and are separate from the 200-scene grounding benchmarks. Tab.~\ref{tab:human-verification} reports the decisions and time cost.

\begin{table}[htbp]
\centering\small
\renewcommand{\arraystretch}{1.12}
\caption{Human verification of newly generated Objects365 masks. SAG and MAG each contribute 500 masks to this audit. Decision rates are percentages; review time is seconds per mask. From-scratch annotation takes approximately 90 seconds per mask.}
\label{tab:human-verification}
\begin{adjustbox}{max width=\linewidth}
\begin{tabular}{@{}lccccc@{}}
\toprule
Regime & Masks & Accept (\%) & Correct (\%) & Reject (\%) & Review (s) \\
\midrule
SAG & 500 & 76.6 & 21.2 & 2.2 & 4.5 \\
MAG & 500 & 65.4 & 30.2 & 4.4 & 10.4 \\
\bottomrule
\end{tabular}
\end{adjustbox}
\end{table}

For SAG, 76.6\% of masks are accepted directly, 21.2\% are corrected, and 2.2\% are rejected, with 4.5 seconds of review per mask. MAG requires more intervention: 65.4\% are accepted, 30.2\% corrected, and 4.4\% rejected, with 10.4 seconds of review per mask. Compared with approximately 90 seconds for from-scratch annotation, these timings indicate a substantial reduction in manual effort through verification and targeted edits. They describe human review time, not end-to-end generation latency or robot control frequency.

\begin{table}[htbp]
\centering\small
\renewcommand{\arraystretch}{1.12}
\caption{MAG branch usage. Percentages are computed over all target instances processed by MAG, including instances without a matched SAM3 mask. They sum to 100\%.}
\label{tab:mag-routing}
\begin{adjustbox}{max width=\linewidth}
\begin{tabular}{@{}lcc@{}}
\toprule
Branch & Operation & Target instances (\%) \\
\midrule
Direct acceptance & Retain a confident SAM3 mask & 53.0 \\
SAG-refine & Refine a low-confidence seed mask & 13.8 \\
SAG-from-scratch & Initialize from an empty mask & 33.2 \\
\bottomrule
\end{tabular}
\end{adjustbox}
\end{table}

Tab.~\ref{tab:mag-routing} reports how the pipeline allocates refinement. Direct acceptance handles 53.0\% of target instances, SAG-refine handles 13.8\%, and SAG-from-scratch handles 33.2\%. The denominator includes all target instances, including those for which SAM3 supplies no matched mask. Thus, the two iterative branches together handle 47.0\% of instances. The multi-instance audit recall is 78.2\% over valid regions, counting corresponding masks regardless of their quality. This coverage-oriented audit differs from Set-R, which requires one-to-one matching at IoU at least 0.5; the two recall values should not be interpreted as the same metric.

\textbf{What refinement contributes.}
Direct acceptance preserves confident SAM3 predictions without invoking the click policy. SAG-refine corrects a supplied candidate while retaining its useful initial geometry; SAG-from-scratch provides a path for targets without an initial mask. These branch frequencies measure how often each operation is used, not the accuracy gain of individual corrected masks. The component ablations in Tab.~\ref{tab:ag-ablation} provide the corresponding pipeline-level evidence: disabling iterative clicks costs 10.4 gIoU points, removing the direct-accept branch costs 4.7, and discarding the refinement seed costs 4.3. Removing dilated gray mask-out costs 2.5 points, and reducing the click cap from eight to two costs 3.2. These are separate ablations of the full system, so their effects are not added together.

\textbf{Automatic refinement versus human correction.}
The 13.8\% SAG-refine share describes automatic inference before human review; the 30.2\% MAG correction rate describes a later human decision over reviewed output masks. Their denominators and stages differ, so they do not form a joint breakdown. The same distinction separates automatic direct acceptance from a reviewer's final acceptance. A mask can enter the review tool after any automatic branch. Rejected masks can subsequently be manually reannotated for additional supervision, but this possibility does not change the reported accept/correct/reject audit outcomes.

\begin{figure}[htbp]
\centering
\includegraphics[width=\linewidth]{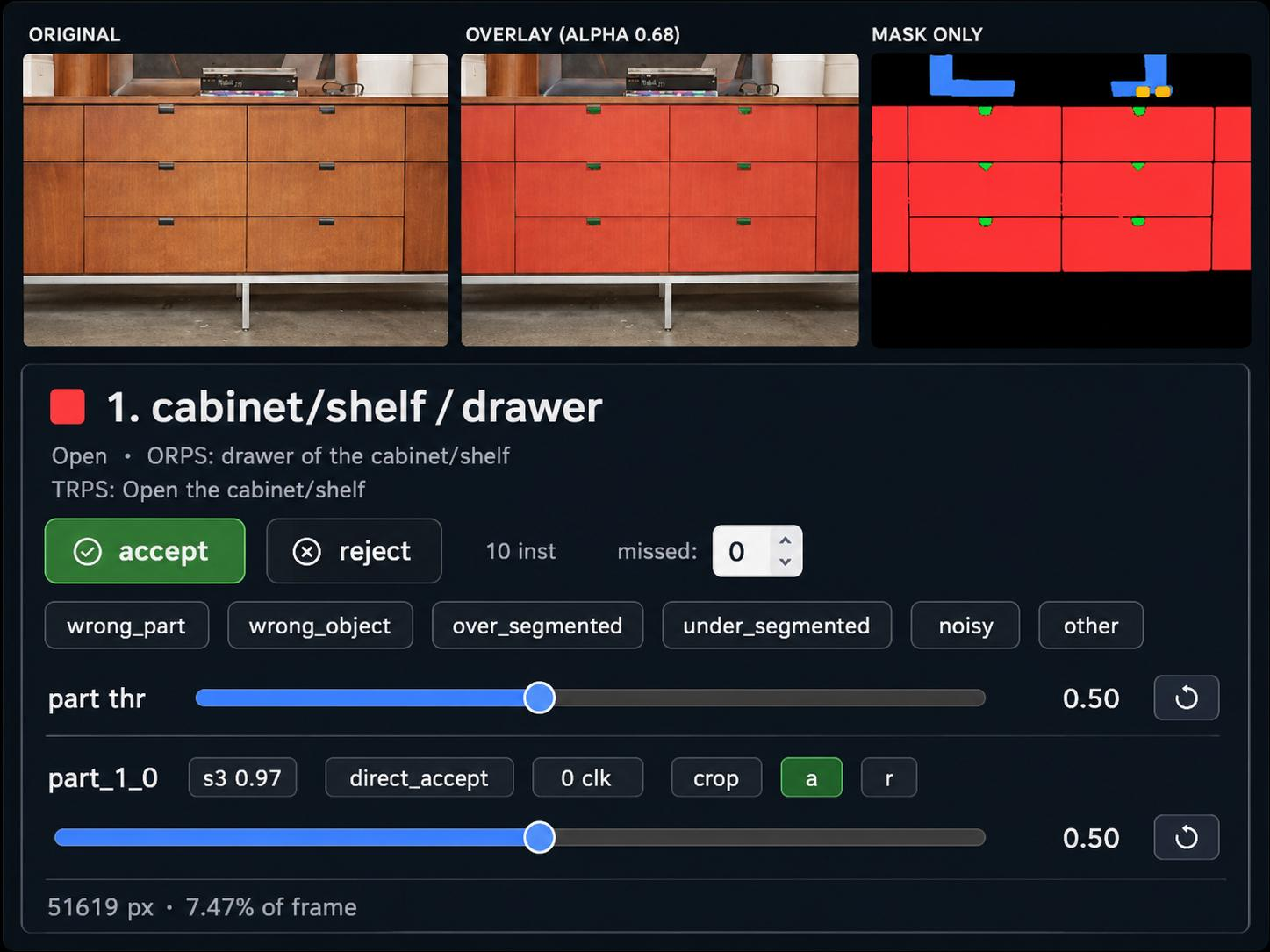}
\caption{Human verification interface. Original, overlay, and mask-only views accompany part-level decisions, threshold controls, branch metadata, and missed-instance counting. The displayed example illustrates the tool; audit statistics are reported in Tab.~\ref{tab:human-verification}.}
\label{fig:annotation-review-ui}
\end{figure}

\textbf{Verification interface.}
Fig.~\ref{fig:annotation-review-ui} shows the original image, mask overlay, and mask-only views used to inspect part annotations. The interface exposes query text, per-part thresholds, automatic branch and click information, missed-instance counts, and accept/reject controls. Reviewers can distinguish wrong objects or parts from over-segmentation, under-segmentation, and noisy boundaries. Per-part click correction and threshold adjustment provide targeted edits, while missed-region counting records omissions that would be invisible if reviewers inspected only the proposed masks. The threshold control displayed in this interface is a review tool setting, not the UAD benchmark threshold.

\FloatBarrier
\section{A2A-Policy: Architecture, Training, and Deployment}
\label{app:policy-implementation}

The comparison changes the affordance interface while keeping the action head and multimodal fusion architecture shared. Both A2A variants use the same frozen A2A-GroundingModel checkpoint in simulation and on the robot. A2A-AffordGen is used for annotation and does not supply masks or features inside the policy loop. Grounding-model parameters remain frozen while robot demonstrations train the policy modules.

\subsection{Visual and Language Interfaces}

\textbf{DP-RGB.} The RGB baseline encodes raw observations with frozen DINOv2-ViT-L/14 features. It receives no predicted affordance overlay. Its Transformer fusion over image, language, and state tokens makes it a strong RGB reference rather than the original 1D-UNet Diffusion Policy architecture.

\textbf{UAD-DP.} The UAD heatmap is thresholded at 0.7 and overlaid on RGB before the same DINOv2 encoder. Overlay observations can be precomputed for the training replay buffer. The policy-overlay threshold is separate from the threshold used to evaluate UAD segmentation accuracy.

\textbf{A2A-Explicit.} A2A-GroundingModel predicts the task-relevant region. The highest-confidence region is thresholded and rendered as a colored alpha overlay before DINOv2 feature extraction. During training, these overlays can be precomputed from the frozen model; at execution, the grounding input is the current observation and instruction.

\textbf{A2A-Implicit.} The same frozen grounding checkpoint supplies multi-scale FPN features. Following the updated interface, the highest-confidence predicted region is selected and thresholded before spatial mask pooling. The region-conditioned features are projected to the policy width and fused with instruction and state tokens. Only the $\langle\mathrm{ACT}\rangle$ token conditions the diffusion head. This is the interface associated with the current 92.8\% LIBERO and 25/40 real-robot aggregates; unfiltered coarse pooling is not used for these results.

\subsection{Shared Architecture and Optimization}

Tab.~\ref{tab:policy-implementation} records the shared architecture and training recipe. The action head predicts a normalized 16-step action chunk from two observation frames, with seven values per action. All four variants train 55.8M parameters in the policy modules while leaving their visual backbones frozen. The current configuration uses per-GPU batch size 16 for all four variants.

\begin{table}[htbp]
\centering\small
\caption{A2A-Policy architecture and LIBERO training settings. The visual/affordance interface differs between policies; fusion and action prediction are shared.}
\label{tab:policy-implementation}
\begin{tabular}{@{}ll@{}}
\toprule
Setting & Value \\
\midrule
Observation history / action horizon & 2 frames / 16 steps \\
Action dimension & 7 \\
Fusion layers / width & 4 / 1024 \\
Attention heads / feed-forward width & 8 / 2048 \\
Diffusion-head depth / width & 6 / 768 \\
Trainable policy parameters & 55.8M \\
Training diffusion steps & 100 \\
Optimizer / learning rate & AdamW / $10^{-4}$ \\
Training epochs / per-GPU batch size & 30 / 16 \\
Warm-up / schedule & 500 steps / cosine decay \\
Weight decay / gradient clipping & $10^{-2}$ / 3.0 \\
Precision / validation interval & \texttt{bf16} / 200 steps \\
DINOv2 input & $448\times448$ \\
Implicit input: simulation / robot & $336\times336$ / $448\times448$ \\
Policy-overlay threshold & 0.7 \\
LIBERO episode cap & 400 steps \\
Replanning interval / DDIM steps & 8 / 8 \\
\bottomrule
\end{tabular}
\end{table}

Training uses AdamW, a warm-up followed by cosine learning-rate decay, and gradient clipping. The 100 training diffusion steps and eight DDIM inference steps refer to different stages of the diffusion model and should not be read as competing execution settings. Both simulation and robot policies replan after executing eight steps of an action chunk.

\subsection{Simulation and Robot Execution}

LIBERO-Object uses 100 expert demonstrations and 50 evaluation episodes per task, with a 400-step cap. Each task requires picking up its named object and placing it in the basket. The aggregate is the mean success rate across the ten tasks; App.~\ref{app:policy-breakdowns} provides the available task-level counts and their corresponding means.

For real-world evaluation, an AgileX Piper six-DoF arm uses head- and wrist-mounted RGB cameras. We collect 100 teleoperated demonstrations per task and train one multi-task policy for each baseline over all four tasks. The RGB front ends use DINOv2-ViT-L/14 with registers at $448\times448$. Language features are provided by frozen MiniLM-L6-v2 with dimension 384. The implicit front end uses the frozen grounding model at $448\times448$, with the same top-1 region selection and thresholded mask pooling as in simulation.

A ZMQ multi-task action server communicates with the Piper client and replans every eight steps using eight DDIM sampling steps. Each task is evaluated with ten trials, so the reported real-robot aggregate is a success count out of 40. The three-seed averages used for grounding accuracy are not applied to these robot trial counts. The qualitative sequences in App.~\ref{app:robot-demos} illustrate execution; the full outcomes are represented by the trial totals.

\section{Per-task Policy Results}
\label{app:policy-breakdowns}

\textbf{Object-wise variation.}
Tab.~\ref{tab:policy-per-task} gives the available per-task breakdown for DP-RGB, UAD-DP, and A2A-Explicit. Its column means match the aggregate results in Tab.~\ref{tab:policy-eval}. For A2A-Implicit, only aggregate success rates are reported.

\begin{table}[htbp]
\centering\small
\renewcommand{\arraystretch}{1.12}
\caption{Per-task LIBERO-Object success rates (\%) for the RGB baseline and the two explicit-conditioning policies, with 50 episodes per task. Every task picks up the named object and places it in the basket. A2A-Implicit is reported at the aggregate level in Tab.~\ref{tab:policy-eval}.}
\label{tab:policy-per-task}
\begin{adjustbox}{max width=\linewidth}
\begin{tabular}{@{}lccc@{}}
\toprule
Target object & DP-RGB & UAD-DP & A2A-Explicit \\
\midrule
Alphabet soup & 88.0 & 96.0 & 96.0 \\
Cream cheese & 90.0 & 50.0 & 98.0 \\
Salad dressing & 92.0 & 100.0 & 100.0 \\
Barbecue sauce & 72.0 & 48.0 & 86.0 \\
Ketchup & 88.0 & 42.0 & 96.0 \\
Tomato sauce & 84.0 & 86.0 & 88.0 \\
Butter & 88.0 & 92.0 & 90.0 \\
Milk & 92.0 & 52.0 & 92.0 \\
Chocolate pudding & 88.0 & 80.0 & 100.0 \\
Orange juice & 96.0 & 98.0 & 98.0 \\
Mean & 87.8 & 74.4 & 94.4 \\
\bottomrule
\end{tabular}
\end{adjustbox}
\end{table}

Explicit conditioning improves cream cheese from 90\% to 98\%, barbecue sauce from 72\% to 86\%, and chocolate pudding from 88\% to 100\%. Milk remains at 92\%, showing that the gain is not uniform across objects. UAD-DP is competitive on some tasks, such as salad dressing, but substantially weaker on others; its aggregate should not be read as a failure on every object. These differences motivate reporting task-level scores alongside the mean.

\textbf{Real-robot task breakdown.}
Tab.~\ref{tab:robot-per-task} reports ten trials for each of the four tasks. Summing each column gives the corresponding total in Tab.~\ref{tab:policy-eval}. The updated A2A-Implicit aggregate is 25/40; its per-task counts are not available, so no task-level allocation is inferred.

\begin{table}[htbp]
\centering\small
\renewcommand{\arraystretch}{1.12}
\caption{Real-robot success counts for the RGB baseline and explicit-conditioning policies. Each task has ten trials. The aggregate A2A-Implicit result is 25/40 (Tab.~\ref{tab:policy-eval}).}
\label{tab:robot-per-task}
\begin{adjustbox}{max width=\linewidth}
\begin{tabular}{@{}lccc@{}}
\toprule
Task & DP-RGB & UAD-DP & A2A-Explicit \\
\midrule
Open the microwave oven & 8/10 & 7/10 & 9/10 \\
Stack the blue cube on the wooden cube & 6/10 & 4/10 & 8/10 \\
Place the cup on the mat & 6/10 & 4/10 & 7/10 \\
Place the phone on the phone stand & 2/10 & 1/10 & 2/10 \\
Total & 22/40 & 16/40 & 26/40 \\
\bottomrule
\end{tabular}
\end{adjustbox}
\end{table}

\FloatBarrier
\section{Real-world Demonstration Sequences}
\label{app:robot-demos}

The four sequences show A2A-Implicit, A2A-Explicit, and UAD-DP from top to bottom. Each figure includes its original Start-to-End arrow indicating left-to-right progression. These qualitative examples supplement the trial counts and do not define a separate evaluation set.


\begin{figure}[htbp]
\centering
\includegraphics[width=0.98\linewidth]{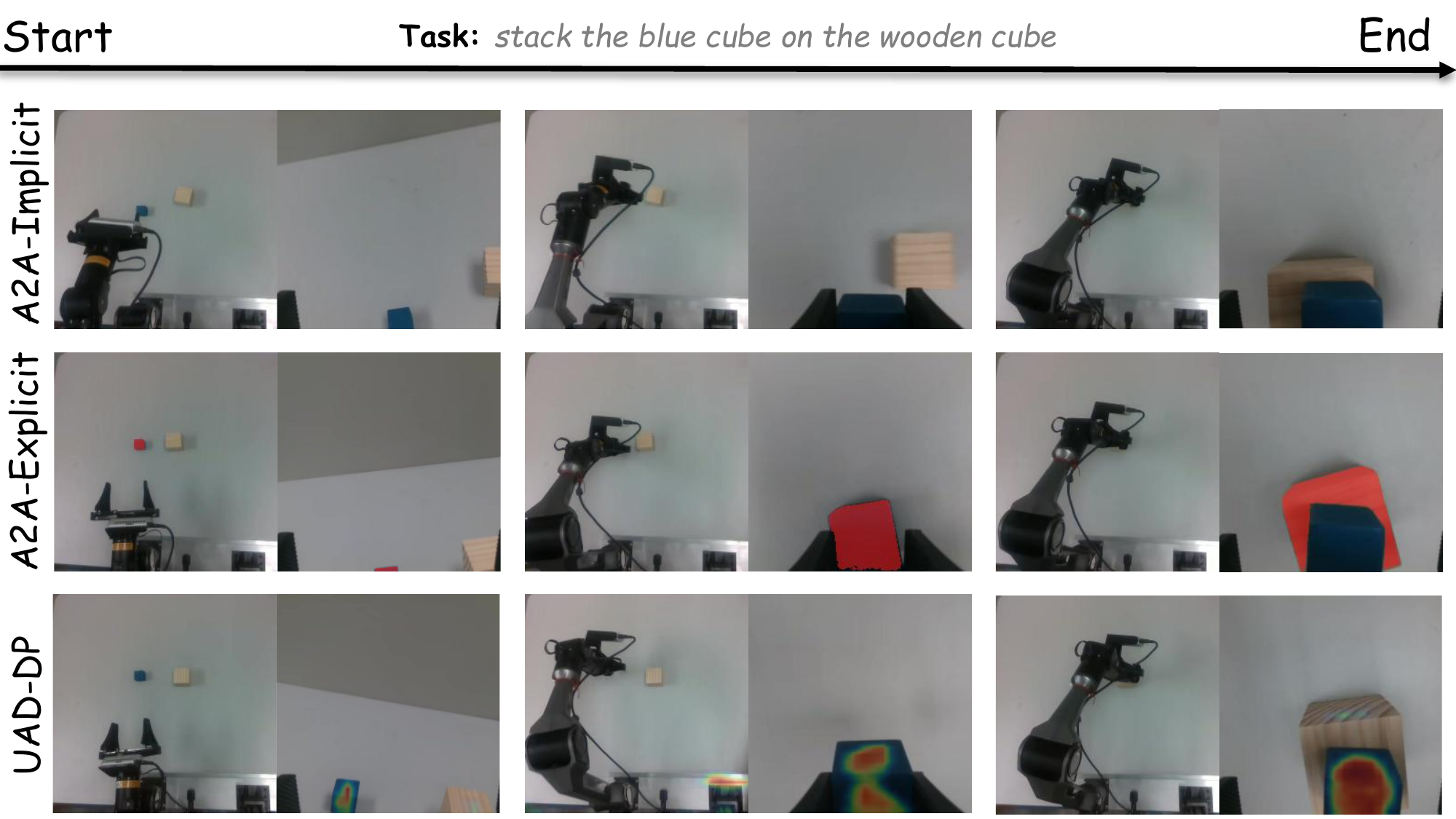}
\caption{Stack the blue cube on the wooden cube. Rows show A2A-Implicit, A2A-Explicit, and UAD-DP from top to bottom; the embedded Start-to-End arrow indicates temporal progression.}
\label{fig:realworld-demo-stack}
\label{fig:realworld-demos}
\end{figure}

\begin{figure}[htbp]
\centering
\includegraphics[width=0.98\linewidth]{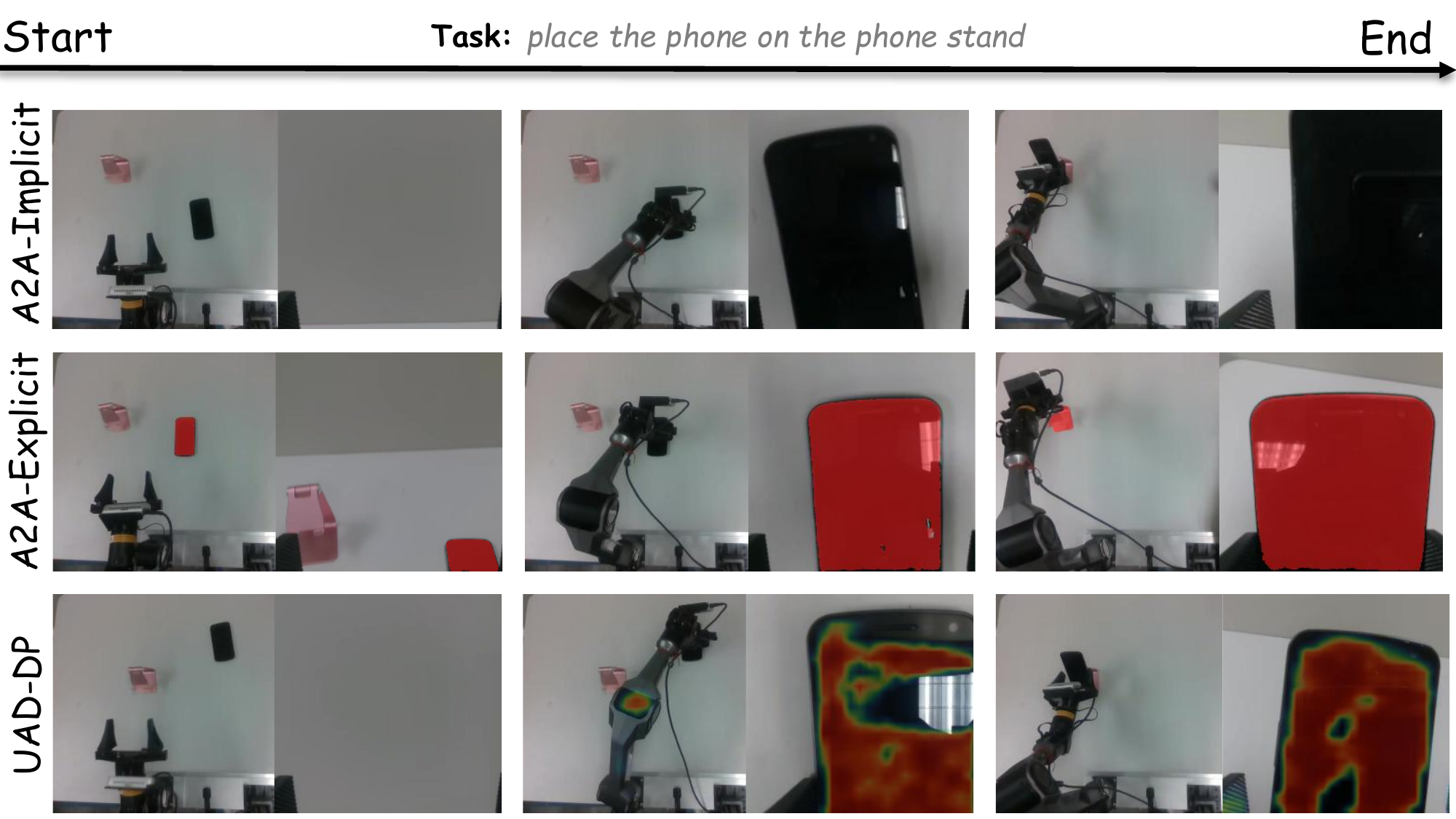}
\caption{Place the phone on the phone stand. Rows show A2A-Implicit, A2A-Explicit, and UAD-DP from top to bottom; the embedded Start-to-End arrow indicates temporal progression.}
\label{fig:realworld-demo-phone}
\end{figure}

\begin{figure}[htbp]
\centering
\includegraphics[width=0.98\linewidth]{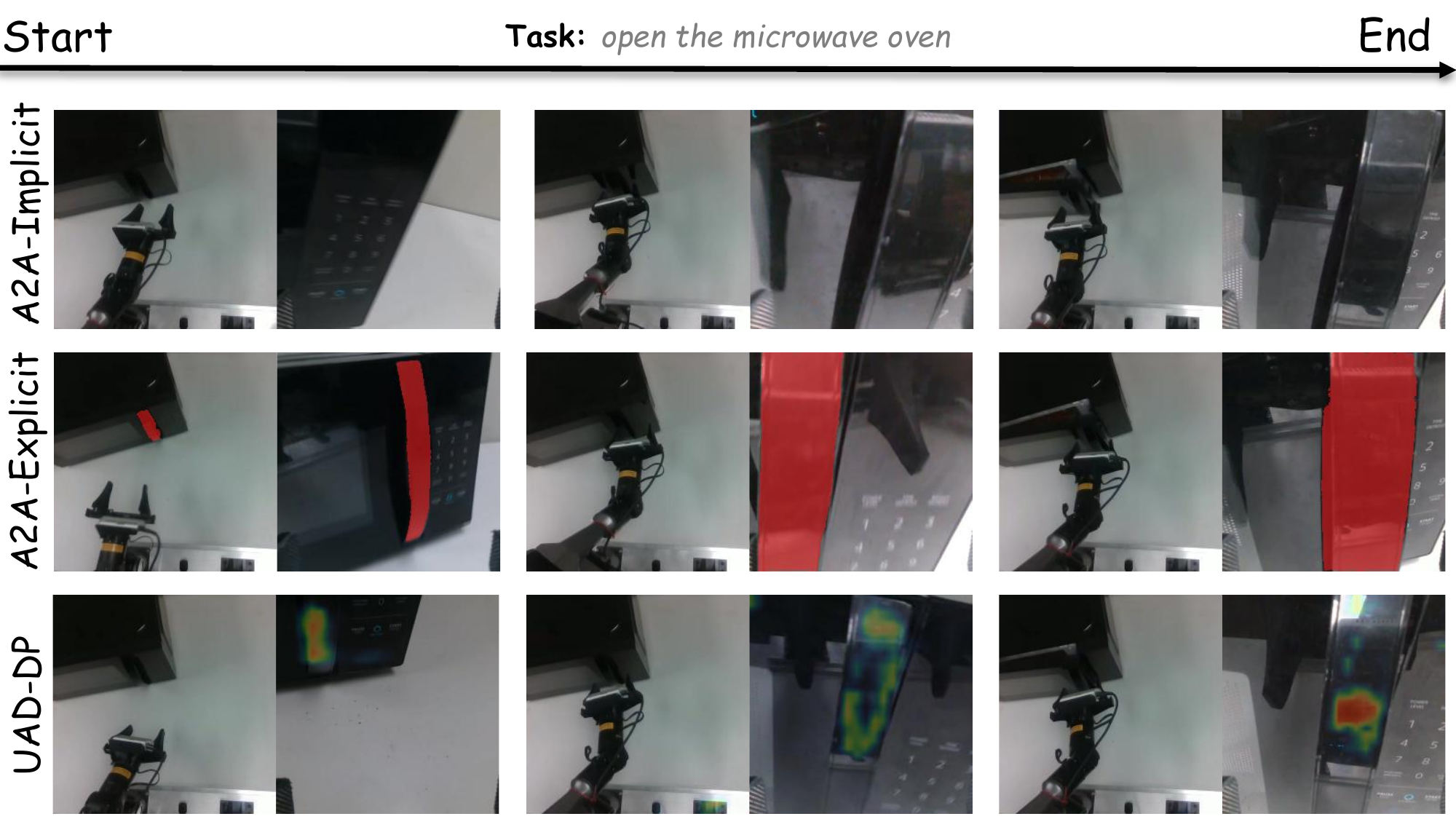}
\caption{Open the microwave oven. Rows show A2A-Implicit, A2A-Explicit, and UAD-DP from top to bottom; the embedded Start-to-End arrow indicates temporal progression.}
\label{fig:realworld-demo-microwave}
\end{figure}

\begin{figure}[htbp]
\centering
\includegraphics[width=0.98\linewidth]{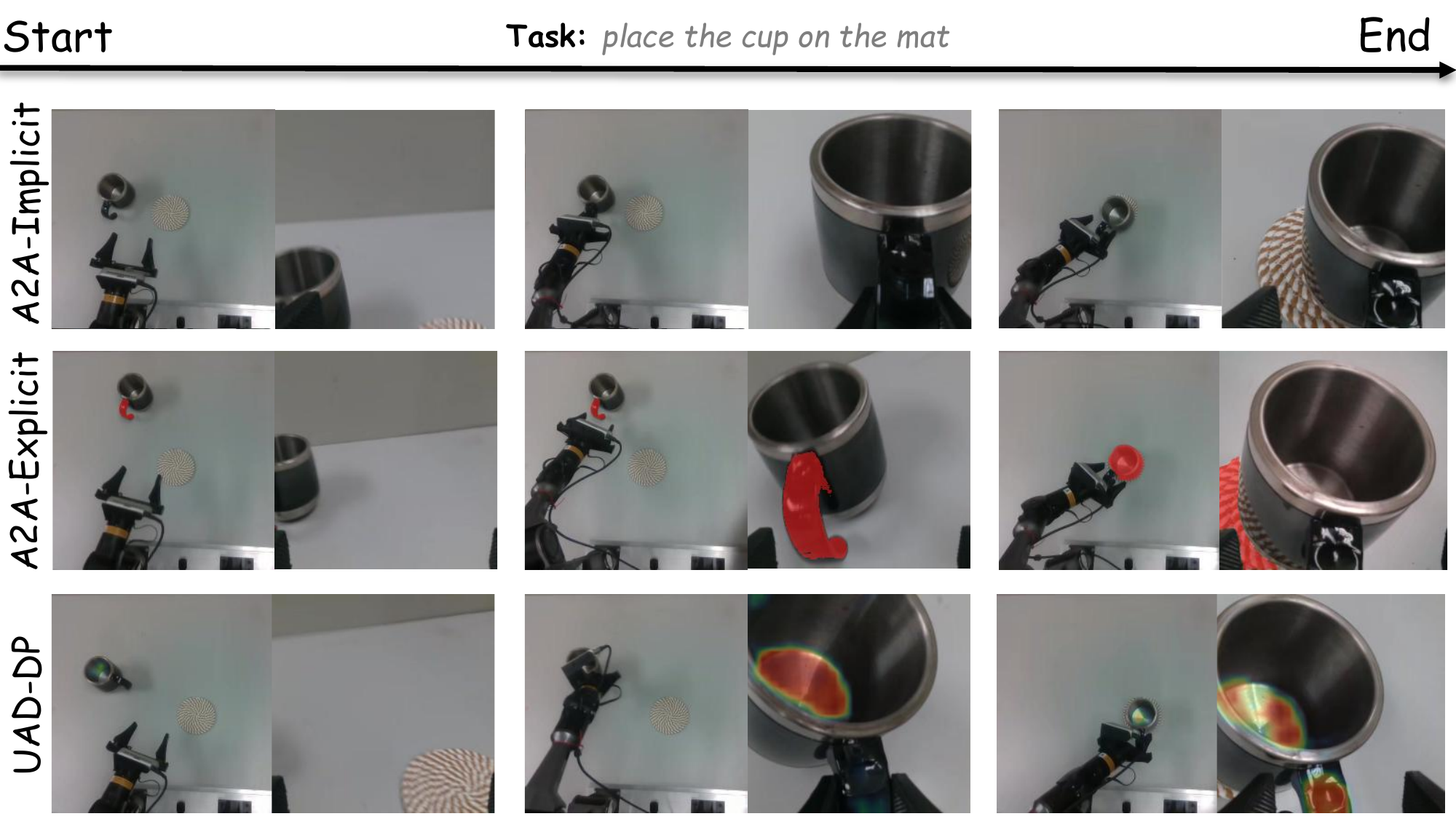}
\caption{Place the cup on the mat. Rows show A2A-Implicit, A2A-Explicit, and UAD-DP from top to bottom; the embedded Start-to-End arrow indicates temporal progression.}
\label{fig:realworld-demo-cup}
\end{figure}
\FloatBarrier
\end{document}